\RequirePackage{fix-cm}
\documentclass[smallcondensed]{svjour3}     
\smartqed  
\usepackage{graphicx}
\usepackage{bbm}
\usepackage{booktabs}
\usepackage[inline]{enumitem}
\usepackage{graphicx}
\usepackage{hyperref}
\usepackage{mathtools}
\usepackage{multirow}
\usepackage[numbers]{natbib}
\usepackage{stmaryrd}
\usepackage[caption=false]{subfig}
\usepackage{tabularx}
\usepackage{textcomp}
\usepackage{verbatim}
\usepackage{xcolor}

\newcolumntype{Y}{>{\centering\arraybackslash}X}

\journalname{Data Mining and Knowledge Discovery}

\begin{document}

\title{Controlling Hallucinations at Word Level in Data-to-Text Generation

}

\author{
    Clement Rebuffel\footnote{Equal contribution} \and
    Marco Roberti\footnotemark[1] \and
    Laure Soulier \and
    Geoffrey Scoutheeten \and
    Rossella Cancelliere \and
    Patrick Gallinari
}

\authorrunning{C. Rebuffel, M. Roberti, L. Soulier, G. Scoutheeten, R. Cancelliere, P. Gallinari}

\institute{Clément Rebuffel \at
           Sorbonne Université, CNRS, LIP6, F-75005 Paris, France \\
              \email{clement.rebuffel@lip6.fr}
           \and
           Marco Roberti \at
              University of Turin, Italy \\
              \email{m.roberti@unito.it}
           \and
           Laure Soulier \at
           Sorbonne Université, CNRS, LIP6, F-75005 Paris, France \\
           \and
           Geoffrey Scoutheeten \at
              BNP Paribas, Paris \\
           \and
           Rossella Cancelliere \at
              University of Turin, Italy \\
           \and
           Patrick Gallinari \at
           Sorbonne Université, CNRS, LIP6, F-75005 Paris, France \\
           Criteo AI Lab, Paris \\
}

\date{Received: date / Accepted: date}
\maketitle

\begin{abstract}
Data-to-Text Generation (DTG) is a subfield of Natural Language Generation aiming at transcribing structured data in natural language descriptions. The field has been recently boosted by the use of neural-based generators which exhibit on one side great syntactic skills without the need of hand-crafted pipelines; on the other side, the quality of the generated text reflects the quality of the training data, which in realistic settings only offer imperfectly aligned structure-text pairs. Consequently, state-of-art neural models include misleading statements --usually called hallucinations-- in their outputs. The control of this phenomenon is today a major challenge for DTG, and is the problem addressed in the paper.

Previous work deal with this issue at the instance level: using an alignment score
for each table-reference pair.
In contrast, we propose a finer-grained approach, arguing that hallucinations should rather be treated at the word level.
Specifically, we propose a Multi-Branch Decoder which is able to leverage word-level labels to learn the relevant parts of each training instance. These labels are obtained following a simple and efficient scoring procedure based on co-occurrence analysis and dependency parsing.
Extensive evaluations, via automated metrics and human judgment on the standard WikiBio benchmark, show the accuracy of our alignment labels and the effectiveness of the proposed Multi-Branch Decoder.

Our model is able to reduce and control hallucinations, while keeping fluency and coherence in generated texts.
Further experiments on a degraded version of ToTTo show that our model could be successfully used on very noisy settings.

\end{abstract}

\section{Introduction}
\label{sec:introduction}

Data-to-Text Generation (DTG) is the subfield of Computational Linguistics and Natural Language Generation (NLG) that is concerned with transcribing structured data into natural language descriptions, or, said otherwise, transcribing machine understandable information into a human understandable description \citep{Gatt2018}. DTG objectives includes \textit{coverage}, i.e.~all the required information should be present in the text, and \textit{adequacy}, i.e. the text should not contain information that is not covered by the input data.
DTG is a domain distinct from other NLG task (e.g.~machine translation \citep{Wiseman2017}, text summarization \citep{Kryscinski2019}) with its own challenges \citep{Wiseman2017}, starting with the nature of inputs \citep{Reiter1997,gardent2020-book}. Such inputs include and are not limited to: databases of records, spreadsheets, knowledge bases, sensor readings. As an example, Fig.~\ref{fig:ex_input} shows an instance of the WikiBio dataset, i.e. a data table containing information about Kian Emadi, paired with its corresponding natural language description found on Wikipedia.

Early approaches to DTG relied on static rules hand-crafted by experts, in which content selection (what to say) and surface realization (how to say it) are typically two separate tasks \citep{Reiter1997,Ferreira2019}. In recent years, neural models have blurred this distinction: various approaches
showed that both content selection and surface realization can be learned in an end-to-end, data-driven fashion \citep{Mei2016,Liu2019hier,Roberti2019, Puduppully2019selplan}. Based on the now-standard encoder-decoder architecture, with attention and copy mechanisms \citep{Bahdanau2014,See2017}, neural methods for DTG are able to produce fluent text conditioned on structured data in a number of domains \citep{Lebret2016,Wiseman2017,Puduppully2019entity}, without relying on heavy manual work from field experts.

Such advances have gone hand in hand with the introduction of larger and more complex benchmarks. In particular, surface-realization abilities have been well studied on hand-crafted datasets such as E2E \citep{Novikova2017e2e} and WebNLG \citep{Gardent2017}, while content-selection has been addressed by automatically constructed dataset such as WikiBio \citep{Lebret2016} or RotoWire \citep{Wiseman2017}.
These large corpora are often constructed from internet sources, which, while easy to access and aggregate, do not consist of perfectly aligned source-target pairs \citep{PerezBeltrachini2017benchmark,Dhingra2019}.
Consequently, model outputs are often subject to over-generation: misaligned fragments from training instances, namely \textit{divergences}, can induce similarly misaligned outputs during inference, the so-called \textit{hallucinations}.

In this paper, we specifically address the issue of hallucinations, which is currently regarded as a major issue in DTG~\citep{gardent2020-book}. Indeed, experimental surveys show that real-life end-users of DTG systems care more about reliability than about readability \citep{reiter2009-investigation}, as unfaithful texts can potentially mislead decision makers, with dire consequences. Hallucinations-reduction methods such as the one presented here have applications in a broad range of tasks requiring high reliability, like news reports~\citep{Leppanen2017}, in which hallucinations may give rise to \textit{fake news}, or summaries of patient information in clinical contexts~\citep{Portet2009,Banaee2013}.

\begin{figure*}
    \centering
    \ttfamily
    \footnotesize
    \begin{tabular}{p{.4\textwidth}l}
    \toprule
    \textsc{key}    & \textsc{value}    \\
    \midrule
    name            & kian emadi        \\
    fullname        & kian emadi-coffin \\
    currentteam     & retired           \\
    discipline      & track             \\
    role            & rider             \\
    ridertype       & sprinter          \\
    proyears        & 2012-present      \\
    proteams        & sky track cycling \\
    \bottomrule
    \\
    \multicolumn{2}{p{.8\textwidth}}{Ref.: kian emadi (born 29 july 1992) is a british track cyclist .}
    \end{tabular}
    \caption{An example of a WikiBio instance, composed by an input table
    and its (partially aligned) description.
    }
    \vspace{-0.5cm}
    \label{fig:ex_input}

\end{figure*}

When corpora include a mild amount of noise, as in handcrafted ones (e.g. E2E, WebNLG), dataset regularization techniques \citep{Nie2019,Dusek2019} or hand crafted rules \citep{Juraska2018} can help to reduce hallucinations.
Unfortunately, these techniques are not suited to more realistic and noisier datasets, as for instance WikiBio \citep{Lebret2016} or RotoWire \citep{Wiseman2017}. On these benchmarks, several techniques have been proposed, such as reconstruction loss terms \citep{Wiseman2017,Wang2019revisiting, Lin2020} or Reinforcement Learning (RL) based methods \citep{PerezBeltrachini2018bootstrap,Liu2019towards,Rebuffel2020rl}. These approaches suffer however from different issues:
\begin{enumerate*}[label=(\arabic*)]
    \item the reconstruction loss relies on the hypothesis of one-to-one alignment between source and target which does not fit with content selection in DTG;
    \item  RL-trained models are based on instance-level rewards (e.g. BLEU \citep{Papineni2002}, PARENT \citep{Dhingra2019}) which can lead to a loss of signal because divergences occur at the word level. In practice, parts of the target sentence express source attributes (in Fig.~\ref{fig:ex_input} name and occupation fields are correctly realized), while others diverge (the birthday and nationality of Kian Emadi are not supported by the source table).
\end{enumerate*}

Interestingly, one can view DTG models as Controlled Text Generation (CTG) ones focused on controlling content, as most CTG techniques condition the generation  on several key-value pairs of \textit{control factors} (e.g.~tone, tense, length) \citep{Dong2017reviews,Hu2017control,Ficler2017control}.
Recently, \citet{Filippova2020hallucinations} explicitly introduced CTG to DTG by leveraging an \textit{hallucination score} simply attached as an additional attribute which reflects the amount of noise in the instance. As an example, the table from Fig~\ref{fig:ex_input} can be augmented with an additional line (\verb|hallucination_score|, \verb|80%|)\footnote{The reader may disagree with such a strong hallucination score. Indeed, while the birthdate and nationality are clearly divergences, the rest of the sentence is correct. This illustrates the complexity of handling divergences in complex datasets, where alignment cannot be framed as a simple word-matching task.}. However, this approach requires a strict alignment at the instance-level, namely between control factors and target text.
A first attempt towards word-level approaches is proposed by \citet{PerezBeltrachini2018bootstrap} (also \textit{PB\&L} in the following). They design word-level alignment labels, denoting the correspondence between the text and the input table, to bootstrap DTG systems. However, they incorporate these labels into a sentence-level RL-reward, which ultimately leads to a loss of this finer-grained signal.\\

In this paper, we go further in this direction with a  DTG model by fully leveraging word-level alignment labels  with a CTG perspective. We propose an original approach in which the word-level is integrated at all phases:
\vspace{-0.2cm}
\begin{itemize}
    \item we propose a \textbf{word-level labeling procedure} (Section~\ref{sec:word-level-hallucination-labels}), based on co-occurrences and sentence structure through dependency parsing. This mitigates the failure of strict word-matching procedure, while still producing relevant labels in complex settings.
    \item we introduce a \textbf{weighted multi-branch neural decoder}(Section~\ref{sec:model}), guided by the proposed alignment labels, acting as word-level control factors. During training, the model is able to distinguish between aligned and unaligned words and learns to generate accurate descriptions without being misled by un-factual reference information.
    Furthermore, our multi-branch weighting approach enables control at inference time.
\end{itemize}
We carry out extensive experiments on WikiBio, to evaluate both our labeling procedure and our decoder (Section~\ref{sec:results}). We also test our framework on ToTTo \citep{Parikh2020}, in which models are trained with  noisy reference texts, and evaluated on references reviewed and cleaned by human annotators to ensure accuracy. Evaluations are based on a range of automated metrics as well as human judgments, and show increased performances regarding hallucinations reduction, while preserving fluency.

Importantly, our approach makes training neural models on noisy datasets possible, without the need to handcraft instances. This work shows the benefit of word-level techniques, which leverage the entire training set, instead of removing problematic training samples, which may form the great majority of the available data.

\section{Related work}
\label{sec:related}

\textbf{Handling hallucinations in noisy datasets.} The use of Deep Learning based methods to solve DTG tasks has led to sudden improvements in state of the art performances \citep{Lebret2016,Wiseman2017,Liu2018,Puduppully2019selplan}.
As a key aspect in determining a model's performance is the quality of training data, several large corpora have been introduced to train and evaluate models' abilities on diverse tasks.
E2E \citep{Novikova2017e2e} evaluates surface realization,  i.e.~the strict transcription of input attributes into natural language; RotoWire \citep{Wiseman2017}
pairs statistics of basketball games with their journalistic descriptions, while
WikiBio \citep{Lebret2016}
maps a Wikipedia info-box with the first paragraph of its associated article. Contrary to E2E, the latter datasets are not limited to surface realization. They were not constructed by human annotators, but rather created from Internet sources, and consist of loosely aligned table-reference
pairs: in WikiBio, almost two thirds of the training instances contain divergences \citep{Dhingra2019}, and no instance has a 1-to-1 source-target alignment \citep{PerezBeltrachini2017benchmark}.

On datasets with a moderate amount of noise, such as E2E, data pre-processing has proven effective for reducing hallucinations. Indeed, rule-based \citep{Dusek2019} or neural-based methods \citep{Nie2019} have been proposed, specifically with table regularization techniques, where attributes are added or removed to re-align table and target description. Several successful attempts have also been made in automatically learning alignments between the source tables and reference texts, benefiting from the regularity of the examples \citep{Juraska2018,Shen2020,Gehrmann2018}. For instance, \citet{Juraska2018} leverage templating and hand-crafted rules to re-rank the top outputs of a model decoding via beam search; \citet{Gehrmann2018} also leverage the
possible templating formats of E2E's reference texts, and train an ensemble of decoders where each decoder is associated to one template; and \citet{Kasner2020} produce template-based lexicalizations and improve them via a \textit{sentence fusion} model.
The previous techniques are not applicable in more complex, general settings. The work of \citet{Dusek2019} hints at this direction, as authors found that neural models trained on E2E were principally prone to omissions rather than hallucinations.
In this direction, \citet{Shen2020} were able to obtain good results at increasing the coverage of neural outputs, by constraining the decoder to focus its attention exclusively on each table cell sequentially until the whole table was realized.
On more complex datasets (e.g.~WikiBio), a wide range of methods has been explored to deal with factualness such as loss design, either with a reconstruction term \citep{Wiseman2017,Wang2019revisiting} or with RL-based methods \citep{PerezBeltrachini2018bootstrap,Liu2019towards,Rebuffel2020rl}. Similarly to the coverage constraints, a reconstruction loss has proven only marginally efficient in these settings, as it contradicts the content selection task \citep{Wang2019revisiting}, and needs to be well calibrated using expert insight in order to bring improvements. Regarding RL, \citet{PerezBeltrachini2018bootstrap} build an instance-level reward which sums up word-level scores; \citet{Liu2019towards} propose a reward based on document frequency to favor words from the source table more than rare words; and \citet{Rebuffel2020rl} train a network with a variant of PARENT \citep{Dhingra2019} using self-critical RL. Note that data regularization techniques have also been proposed \citep{Thomson2020,Wang2019revisiting}, but these methods require heavy manual work and expert insights, and are not readily transposable from one domain to another.\newline

\noindent\textbf{From CTG to controlling hallucinations.}
Controlled Text Generation (CTG) is concerned with constraining a language model's output during inference on a number of desired attributes, or \textit{control factors},
such as the identity of the speaker in a dialog setting \citep{Li2016persona}, the politeness of the generated text or the text length in machine-translation
\citep{Sennrich2016politeness,Kikuchi2016length}, or the tense in generated movie reviews \citep{Hu2017control}. Earlier attempts at neural CTG
can even be seen as direct instances of DTG as it is currently defined:
models are trained to generate text conditioned on attributes of interest, where attributes are key-value pairs. For instance, in the movie review domain, \citet{Ficler2017control} proposed an expertly crafted dataset, where sentences are strictly aligned with control factors, being either content or linguistic style aspects (e.g.~tone, length).\\
\indent In the context of dealing with hallucinations in DTG, \citet{Filippova2020hallucinations} recently proposed a similar framework, by augmenting source tables with an additional attribute that reflects the degree of hallucinated content in the associated target description.
During inference, this attribute acts as an \textit{hallucination handle} used to produce more or less factual text.
As mentioned in Section~\ref{sec:introduction}, we argue that a unique value can not accurately represent the correspondence between a table and its description, due to the phrase-based nature of divergences. \\

Based on the literature review, the lack of model control can be evidenced when loss modification methods are used \citep{Wang2019revisiting,Liu2019hier,Rebuffel2020rl},
although these approaches can be efficient and transposed from one domain to another. On the other hand, while CTG deals with control and enables choosing the defining features of generated texts~\citep{Filippova2020hallucinations}, standard approaches rely on instance-level control factors that do not fit with hallucinations, which rather appear due to divergences at the word level. Our approach aims at gathering the merits of both trends of models and is guided  by previous statements highlighting that word-level is primary in hallucination control. More particularly, our model differs from previous ones in several aspects:

\begin{enumerate}[label=(\arabic*)]
    \item Contrasting with data-driven approaches (i.e. dataset regularization) which are costly in expert time, and loss-driven approaches (i.e. reconstruction or RL losses) which often do not take into account key subtasks of DTG (content-selection, world-level correspondences), we propose a multi-branch modeling procedure which allows the controllability of the hallucination factor in DTG. This multi-branch model can be integrated seamlessly in current approaches, allowing to keep peculiarities of existing DTG models, while deferring hallucination management to a parallel decoding branch.

    \item Unlike previous CTG approaches \citep{Li2016persona,Sennrich2016politeness,Ficler2017control,Filippova2020hallucinations} which propose instance-level control factors, the control of the hallucination factor is performed at the word-level to enable finer-grained signal to be sent to the model.

\end{enumerate}

Our model is composed of two main components:
\begin{enumerate*}[label=(\arabic*)]
    \item a word-level alignment labeling mechanism,  which makes the correspondence between the input table and the text explicit, and
    \item a multi-branch decoder guided by these alignment labels. The branches separately integrate co-dependent control factors (namely content, hallucination and fluency).
\end{enumerate*}
We describe these components in Sections~\ref{sec:word-level-hallucination-labels} and~\ref{sec:model}, respectively.

    \vspace{-0.4cm}
\section{Word-level Alignment Labels}
\label{sec:word-level-hallucination-labels}
    \vspace{-0.2cm}

\begin{figure*}
\centering
    \includegraphics[width=\textwidth]{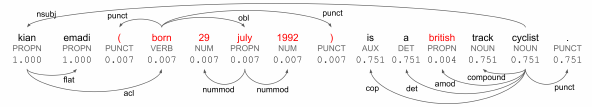}

    \caption{The reference sentence of the example shown in Fig.~\ref{fig:ex_input}. Every token is associated to its Part-of-Speech tag and hallucination score $s_t$. Words in red denote $s_t<\tau$. The dependency parsing is represented by labeled arrows that flow from parents to children. Important words are \textit{kian}, \textit{emadi}, \textit{29}, \textit{july}, \textit{1992}, \textit{british}, \textit{track}, and \textit{cyclist}.}
    \label{fig:ex_output}
    \vspace{-0.4cm}
\end{figure*}

We consider a  DTG task, in which the corpus $\mathcal{C}$ is composed of a set of
entity-description pairs, $(e, y)$. A \textit{single-entity table} $e$ is a variable-sized set of $T_e$ key-value pairs $x \coloneqq (k, v)$.
A \textit{description} $y \coloneqq
y_{1:T_y}$ is a
sequence of $T_y$ tokens representing the natural language description of the
entity; we refer to the tokens spanning from indices $t$ to $t'$ of a description $y$ as $y_{t:t'}$. A description is made of \textit{statements}, defined as text spans expressing one single idea (Appendix~\ref{appendix-sec:alignment-labels} presents in detail the statement partitioning procedure). We refer to the first index of a statement as $t_i$, so that $y_{t_i:t_{i+1}-1}$ is the $i^{th}$ statement itself.
Fig.~\ref{fig:ex_input} shows a WikiBio entity
made by 8 key-value pairs
together with its associated description.

First, we aim at labeling each word from a description, depending on the presence of a correspondence with its associated table. We call such labels \textit{alignment labels}. We drive the word-level labeling procedure on two intuitive constraints:
\begin{enumerate*}[label=(\arabic*)]
    \item important words (names, adjectives and numbers) should be labeled depending on their alignment with the table, and
    \item words from the same statement should have the same label.
\end{enumerate*}

With this in mind, the \textit{alignment label} for the $t^{\text{th}}$ token $y_t$ is a binary label: $l_t \coloneqq \mathbbm{1}_{\{s_t > \tau\}}$ where $s_t$ refers to the \textit{alignment score} between $y_t$ and the table, and $\tau$ is set experimentally (see Sec.~\ref{subsec:implementation-details}).
The \textit{alignment score} $s_t$  acts as a normalized measure of correspondence between a token $y_t$ and the table $e$:
\begin{equation}
    s_t \coloneqq \mathit{norm}(\max_{x \in e}\mathit{align}(y_t, x), ~~y)
    \label{eq:alignment-score}
        \vspace{-0.2cm}
\end{equation}
where the function \textit{align} estimates the alignment between token $y_t$ and a key-value pair $x$ from the input table $e$, and \textit{norm} is a normalization function based on the dependency structure of the description $y$.
Fig.~\ref{fig:ex_output} illustrates our approach: under each word we show its word alignment score, and words are colored in red if this score is lower than $\tau$, denoting an alignment label equal to 0.
Below, we describe these functions
(Appendix~\ref{appendix-sec:alignment-labels} contains reproducibility details).

\vspace{2mm}
\noindent\textbf{Co-occurrence-based alignment function ($\mathbf{align(\cdot, x)}$).}
This function assigns to important words a score in the interval $[0,1]$ proportional to their co-occurrence count (a proxy for alignment) with the key-value pair from the input table.
If the word $y_t$ appears in the key-value pair $x \coloneqq (k,v)$,  $align(y_t, x)$  outputs $1$; otherwise, the output is obtained scaling the number of occurrences $co_{y_t,x}$ between $y_t$ and $x$ through the dataset:
\begin{equation}\small
    align(y_t,x) \coloneqq
    \begin{cases}
    1 & \text{if~~} y_t \in x\\
    a \cdot (co_{y_t,x}\!- m)^2& \text{if~~} m \le co_{y_t,x}\!\le M\\
    0& \text{if~~} 0 \le co_{y_t,x}\!\le m
    \end{cases}
\end{equation}
where $M$ is the maximum number of word co-occurrences in the dataset vocabulary and the row $x$,
$m$ is a threshold value, and $a \coloneqq \frac{1}{(M-m)^2}$.

\vspace{2mm}
\noindent\textbf{Score normalization ($\mathbf{norm(\cdot, y)}$).}
According to the already stated assumption~(2)  -- words inside the same statement should have the same score -- , we first split the sentence $y$ into statements $y_{t_i:t_{i+1}-1}$, via dependency parsing and its rule-based conversion to constituency trees
 \citep{Han2000, Xia2001, Hwa2005, Borensztajn2009}.
Given a word $y_t$ associated to the score $s_t$ and belonging to statement $y_{t_i:t_{i+1}-1}$, its normalized score corresponds to the average score of all important words in this statement:
    \vspace{-0.2cm}
\begin{equation}
    norm(s_t, y) = \frac{1}{t_{i+1}-t_i} \sum_{j=t_i}^{t_{i+1}-1} s_j
        \vspace{-0.2cm}
\end{equation}

This in-statement average depends on both the specific word and its context, leading to coherent hallucination scores which can be thresholded without affecting the syntactical sentence structure, as shown in Fig.~\ref{fig:ex_output}.

    \vspace{-0.4cm}
\section{Multi-Branch Architecture}
\label{sec:model}

The proposed Multi-Branch Decoder (MBD) architecture aims at separating targeted co-dependent factors during generation. We build upon the standard DTG architecture, an encoder-decoder with attention and copy mechanism, which we modify by duplicating the decoder module into three distinct parallel modules. Each control factor (i.e. content, hallucination or fluency) is modeled via a single decoding module, also called branch, whose output representation can be weighted according to its desired importance.
At training time, weights change depending on the word currently being decoded, inducing the desired specialization of each branch. During inference, weights are manually set, according to the desired trade-off between information reliability, sentence diversity and global fluency. Text generation is thus controllable, and consistent with the  control factors.

\begin{figure*}
    \centering
    \includegraphics[width=\textwidth]{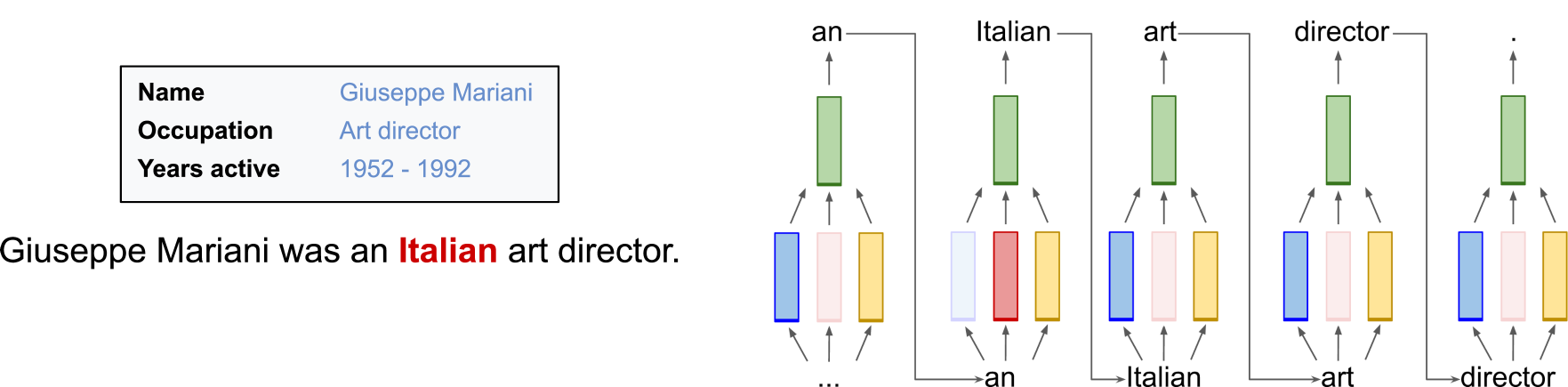}
    \caption{
    Our proposed decoder with three branches associated to content (in blue -- left), hallucination (in red -- middle) and fluency (in yellow -- right).
    Semi-transparent branches are assigned the weight $0$.
    }
    \label{fig:model}
    \vspace{-0.4cm}
\end{figure*}

Figure~\ref{fig:model} illustrates a training step over the sentence ``\textit{Giuseppe Mariani was an Italian art director}'', in which \textit{Italian} is a divergent statement (i.e. is not supported by the source table). While decoding factual words, the weight associated to the content (resp. hallucination) branch is set to $0.5$ (resp. $0$) while during the decoding of \textit{Italian}, the weight associated to the content (resp. hallucination) branch is set to $0$ (resp. $0.5$). Note that the weight associated to the fluency branch is always set to $0.5$, as fluency does not depend on factualness.

The decoding modules' actual architecture may vary, as we framed the MBD model from a high level perspective. Therefore, all types of decoder can be used, such as Recurrent Neural Networks (RNNs)~\citep{Rumelhart1986}, Transformers~\citep{Vaswani2017}, and Convolutional Neural Networks~\citep{Gehring2017}. The framework can be generalized to different merging strategies as well, such as late fusion, in which the final distributions are merged, instead of the presented early fusion, which works at the decoder states level.

In this paper, experiments are carried out on RNN-based decoders, weighting their hidden states. Section~\ref{sub:standard-architecture} presents the standard DTG encoder-decoder architecture; Section~\ref{sub:modeldecoder} shows how it can be extended to MBD, together with its peculiarities and the underlying objectives and assumptions.

    \vspace{-0.4cm}
\subsection{Standard DTG architecture}
\label{sub:standard-architecture}
    \vspace{-0.2cm}

Neural DTG approaches  typically use an encoder-decoder architecture \citep{Wiseman2017} in which
\begin{enumerate*}[label=(\arabic*)]
    \item the encoder relies on a RNN to encode each element of the source table into a fixed-size latent representation $h_j$ (elements of the input table are first embedded into $T_e$ $N$-dimensional vectors, and then fed sequentially to the RNN \citep{Wiseman2017}), and
    \item the decoder  generates a textual description $y$ using a RNN augmented with attention and copy mechanisms \citep{See2017}.
\end{enumerate*}
Words are generated in an auto-regressive way. The decoder's RNN updates its hidden state $d_t$ as:
\begin{equation}
    d_t \coloneqq \text{RNN}(d_{t-1}, [y_{t-1}, c_t])
\end{equation}
where $y_{t-1}$ is the previous word and $c_t$ is the context vector obtained through the attention mechanism. Finally, a word is drawn from the distribution computed via a copy mechanism \citep{See2017}.

    \vspace{-0.2cm}
\subsection{Controlling Hallucinations via a Multi-Branch Model}
\label{sub:modeldecoder}
    \vspace{-0.2cm}

\begin{sloppypar}
Our objective is to enrich the decoder in order to be able to tune the content/hallucination ratio during generation, aiming at enabling generation of hal\-lu\-ci\-na\-tion-free text when needed. Our key assumption is that the decoder's generation is conditioned by three co-dependent factors:
\end{sloppypar}
\begin{itemize}[nosep,left=0em,itemindent=1em]
   \item \textit{Content factor} constrains the generation to realize only the information included in the input;
    \item \textit{Hallucinating factor} favors lexically richer and more diverse text, but may lead to hallucinations not grounded by the input;
    \item \textit{Fluency factor}\footnote{\citet{Wiseman2018} showed that the explicit modeling of a fluency latent factor improves performance.} conditions the generated sentences toward global syntactic correctness, regardless of the relevance.
\end{itemize}

Based on this assumption, we propose a multi-branch encoder-decoder network, whose branches are constrained on the above factors at word-level, as illustrated in Fig.~\ref{fig:model}.
Our network has a single encoder and $F=3$ distinct decoding RNNs, noted $\text{RNN}^f$ respectively, one for each factor.
During each decoding step, the previously decoded word $y_{t-1}$ is fed to all RNNs, and a final decoder state $d_t$ is computed using a weighted sum of all the corresponding hidden states,
\begin{gather}
    d_t^f \coloneqq \text{RNN}^f(d_{t-1}^f, [y_{t-1}, c_t])\\
    d_t \coloneqq \sum_{f=1}^F \omega_t^f d^f_t
\end{gather}

where $d_t^f$ and $\omega_t^f$ are respectively the hidden state and the weight of the $f^{th}$ RNN at time $t$.

Weights are used to constrain the decoder branches to the desired control factors ($\omega_t^0,\omega_t^1,\omega_t^2$ for the content, hallucination and fluency factors respectively) and sum to one.

During training, their values are dynamically set depending on the \textit{alignment label} $l_t\in\{0,1\}$ of the target token $y_t$ (see Sec. \ref{subsec:implementation-details}).
While a number of mappings can be used to set the weights given the alignment label, early experiments have shown that better results were achieved when using a binary switch for each factor, i.e.~activating/deactivating each branch, as shown in Fig.~\ref{fig:model} (note that fluency should not depend on content and therefore its associated branch is always active).

During inference, the weights of the decoder's branches are set manually by a user, according to the desired trade-off between information reliability, sentence diversity and global fluency. Text generation is then controllable and consistent with the  control factors.

    \vspace{-0.1cm}
\section{Experimental setup}
    \vspace{-0.2cm}
\label{sec:experiments}
\subsection{Datasets}
\label{subsec:datasets}
    \vspace{-0.2cm}
We evaluated the model on two representative large size datasets. Both have been collected automatically and present a significant amount of table-text divergences for training. Both datasets involve content selection and surface realization, and represent a relatively realistic setting.

\vspace{2mm}
\noindent\textbf{WikiBio \citep{Lebret2016}}
contains $728,321$ tables, automatically paired with the first sentence of the corresponding Wikipedia English article.
Reference text's average length is $26$ words, and tables have on average 12 key-value pairs. We use the original data partition: $80\%$ for the train set, and $10\%$ for validation and test sets.
This dataset has been automatically built from the Internet;
concerning divergences, $62\%$ of the references mention extra information not grounded by the  table \citep{Dhingra2019}.

\vspace{2mm}
\noindent\textbf{ToTTo \citep{Parikh2020}}
contains $120,761$ training examples, and $7,700$ validation and test examples.
For a given Wikipedia page, an example is built up by pairing its summary table and a candidate sentence, selected across the whole page via simple similarity heuristics. Such a sentence may accordingly realize whichever table cells, making content selection arbitrary; furthermore, its lexical form may strongly depend on the original context, because of pronouns or anaphoras. Divergences are of course present as well. Those issues have been addressed by~\citet{Parikh2020} by
\begin{enumerate*}[label=(\arabic*)]
    \item \textit{highlighting} the input cells realized by the output, and
    \item removing divergences and making the sentence self-contained (e.g.~replacing pronouns with their invoked noun or noun phrase).
\end{enumerate*}
Fig.~\ref{fig:totto-quali-ex} exemplifies the difference between noisy and clean ToTTo sentences. In our experiments, we limit the input to the highlighted cells and use the original, noisy sentence as output. Noisy texts' average length is $17.4$ words, and $3.55$ table cells are highlighted, on average.

    \vspace{-0.4cm}
\subsection{Baselines}
\label{subsec:baselines}
    \vspace{-0.2cm}

We assess the accuracy and relevance of our alignment labels against the ones proposed by \citet{PerezBeltrachini2018bootstrap}, which is, to the best of our knowledge, the only work proposing such a fine-grained alignment labeling.

To evaluate our Multi-Branch Decoder (\textit{MBD}), we consider five baselines:
\begin{itemize}[nosep,left=0em,itemindent=1em]
    \item \textit{stnd} \citep{See2017}, a LSTM-based encoder-decoder model with attention and copy mechanisms. This is the standard sequence-to-sequence recurrent architecture.
    \item \textit{stnd\_filtered}, the previous model trained on a filtered version of the training set: tokens deemed hallucinated according to their hallucination scores, are removed from target sentences.
    \item \textit{hsmm} \citep{Wiseman2018}, an encoder-decoder model with a multi-branch decoder. The branches are not constrained by explicit control factors. This is used as a baseline to show that the multi-branch architecture by itself does not guarantee the absence of hallucinations.
    \item \textit{hier} \citep{Liu2019hier}, a hierarchical sequence-to-sequence model, with a coarse-to-fine attention mechanism to better fit the \textit{attribute-value} structure of the tables. This model is trained with three auxiliary tasks to capture more accurate semantic representations of the tables.
    \item \textit{hal\textsubscript{WO}} \citep{Filippova2020hallucinations}, a \textit{stnd}-like model trained by augmenting each source table with an additional attribute (\textit{hallucination ratio}, \textit{value}).
\end{itemize}

We ran our own implementations of \textit{stnd}, \textit{stnd\_filtered} and \textit{hal\textsubscript{WO}}. Authors of \textit{hier} and \textit{hsmm} models kindly provided us their WikiBio's test set outputs. The metrics described in Sec.~\ref{subsec:metrics} were directly applied on them.

    \vspace{-0.2cm}
\subsection{Implementation Details}
\label{subsec:implementation-details}

During training of our multi-branch decoder the fluency branch is always active ($\omega_t^2 = 0.5$) while the content and hallucination branches are alternatively activated, depending on the alignment label $l_t$:  $\omega_t^0 = 0.5$ (content factor) and $\omega_t^1 = 0$ (hallucination factor) when $l_t = 1$, and conversely. The threshold $\tau$ used to obtain $l_t$ is set to $0.4$ using human tuning to optimize for highest accuracy\footnote{Note that accuracy is not heavily impacted by different choices of $\tau$. We report in Appendix~\ref{appendix-sec:implementation-details} the respective accuracy scores of our proposed automated labels for different values of $\tau$.}.
All hyperparameters were tuned in order to optimize the validation PARENT F-measure \citep{Dhingra2019}. In particular, we use the [$0.4$~$0.1$~$0.5$] weight combination during inference. See Sec.~\ref{subsec:weights-impact} for a discussion about weight combinations and Appendix~\ref{appendix-sec:implementation-details} for other implementation details.\footnote{Code is given to reviewers and will be available upon acceptance.}
    \vspace{-0.4cm}
\subsection{Metrics}
\label{subsec:metrics}
    \vspace{-0.2cm}
To evaluate our model, we carried out
\begin{enumerate*}[label=(\arabic*)]
    \item an automatic analysis
    and
    \item a human evaluation for a qualitative analysis of generated sentences.
\end{enumerate*}

For the automatic analysis, we use five metrics:
\begin{itemize}[nosep,left=0em,itemindent=1em]
    \item BLEU \citep{Papineni2002} is a length-penalized precision score over $n$-grams, $n \in \llbracket 1, 4 \rrbracket$,   optionally improved with a smoothing technique \citep{Chen2014}.
    Despite being the standard choice, recent findings show that it correlates poorly with human evaluation, especially on the sentence level \citep{Novikova2017metrics,Reiter2018}, and that it is a proxy for sentence grammar and fluency aspects rather than semantics \citep{Dhingra2019}.
    \item PARENT \citep{Dhingra2019} computes smoothed $n$-gram precision and recall
    over both the reference and the input table. It
    is explicitly designed for DTG tasks, and its F-measure shows
    ``the highest correlation with humans across a range of settings with divergent references in WikiBio.'' \citep{Dhingra2019}
    \item  The \textit{hallucination rate} computes the percentage of tokens labeled as hallucinations (Sec.~\ref{sec:word-level-hallucination-labels}).
    \item The average generated sentence length in number of words.
    \item The classic readability Flesch index \citep{flesch}, which is based on words per sentence and syllables per word, and is still used as a standard benchmark \citep{Kosmajac2019,Smeuninx2020,Stajner2020a,Stajner2020b}.
\end{itemize}

Finally, we perform qualitative evaluations of the results obtained on WikiBIO and ToTTo, following the best practices outlined by \citet{Lee2019}. Our human annotators are from several countries across Europe, between 20 and 55 years old and proficient in English. They have been assigned two different tasks:
\begin{enumerate*}[label=(\roman*)]
    \item hallucination labeling, i.e.~the selection of
    sentence pieces which include incorrect information, and
    \item sentence analysis, i.e.~evaluating different realizations of the same table according to their fluency, factualness and coverage. Scores are presented as a 3-level Likert scale for Fluency (\textit{Fluent}, \textit{Mostly fluent}, or \textit{Not fluent}) and Factualness (likewise), while coverage is the number of cells from the table that have been realized in the description.
\end{enumerate*}
To avoid all bias, annotators are shown a randomly selected table at a time, together with its corresponding descriptions, both from the dataset and the models that are being evaluated. Sentences are presented each time in a different order.
Following \citet{Tian2019}, we first tasked three expert annotators to annotate a pilot batch of 50 sentences. Once confirmed that Inter-Annotator Agreement was approx. $75$\% (a similar finding to \citet{Tian2019}), we asked 16 annotators to annotate a bigger sample of 300 instances (where each instance consists of one table and four associated outputs), as \citet{Liu2019hier}.\footnote{
An eyesight of our platform is available in Appendix~\ref{sec:interface}.}

    \vspace{-0.4cm}
\section{Results}
\label{sec:results}

We perform an extensive evaluation of our scoring procedure and multi-branch architecture on the WikiBio dataset: we evaluate - the quality of the proposed alignment labels, both intrinsically using human judgment and extrinsically by means of the DTG downstream task and - the performance of our model with respect to the baselines. Additionally, we assess the applicability of our framework on the more noisy ToTTo benchmark, which represents a harder challenge for today's DTG models.

\begin{table}
    \centering
    \begin{tabular}{ccccc}
    \toprule
    Labels & Accuracy & Precision & Recall & F-measure \\
    \midrule
    PB\&L       & 46.9\%   & 21.3\%    & 49.2\% & 29.7\%       \\
    ours       & \bf 87.5\%   & \bf 80.6\%    & \bf 59.8\% & \bf 68.7\%    \\
    \bottomrule
    \toprule
    \multirow{3}*{Labels} & \multirow{3}*{BLEU} & \multicolumn{3}{c}{PARENT}  \\
    \cmidrule(lr){3-5}
    && Precision & Recall & F-measure \\
    \midrule
    PB\&L      & 32.15\%	&  76.91\%  &   39.28\%  &   48.75\%   \\
    ours       & \bf 40.51\%	& \bf 77.71\%  &  \bf 45.01\%  &  \bf 54.57\%   \\
    \bottomrule
    \end{tabular}
    \caption{Performances of hallucination scores on WikiBio test set, w.r.t. human-designated labels (upper table) and  \textit{MBD} trained with different labeling procedures (lower table). Our model always significantly overpasses \textit{PB\&L} (T-test with $p < 0.005$). }

    \label{tab:scorecompare}
        \vspace{-0.5cm}
\end{table}

    \vspace{-0.4cm}
\subsection{Validation of Alignment Labels.}
\label{subsec:validation-alignment-labels}
    \vspace{-0.2cm}

To assess the effectiveness of our alignment labels (Sec.~\ref{sec:word-level-hallucination-labels}), we first compare the alignment labels against human judgment, and then explore their impact on a DTG  task. As a baseline for comparison we report performances of \textit{PB\&L}.

\begin{figure*}
	\centering
	\scriptsize
    \subfloat[][\label{tab:scoring_quali1}]{
	\ttfamily
    \begin{tabularx}{\linewidth}{lX}

		\toprule
		\textsc{key}   & \textsc{value} \\
		\midrule
		name           & patricia flores fuentes               \\
		birth\_date    & 25 july 1977 \\
		birth\_place   & state of mexico , mexico              \\
		occupation     & politician \\
		nationality    & mexican \\
		article\_title & patricia flores fuentes               \\
		\bottomrule
		\\
		\multicolumn{2}{p{.95\linewidth}}{\textbf{Ref.}: patricia flores fuentes -lrb- born 25 july 1977 -rrb- is a mexican politician affiliated to the national action party .}\\
		\multicolumn{2}{p{.95\linewidth}}{\textbf{PB\&L}: patricia flores fuentes \textcolor{red}{\underline{-lrb-}} \textcolor{red}{\underline{born}} 25 july 1977 \textcolor{red}{\underline{-rrb-}} \textcolor{red}{\underline{is}} \textcolor{red}{\underline{a}} mexican politician \textcolor{red}{\underline{affiliated}} \textcolor{red}{\underline{to}} \textcolor{red}{\underline{the}} \textcolor{red}{\underline{national}}
		\textcolor{red}{\underline{action}} party \textcolor{red}{\underline{.}}}\\
		\multicolumn{2}{p{.95\linewidth}}{\textbf{Ours}: patricia flores fuentes -lrb- born 25 july 1977 -rrb- is a mexican politician \textcolor{red}{\underline{affiliated}} \textcolor{red}{\underline{to}} \textcolor{red}{\underline{the}} \textcolor{red}{\underline{national}} \textcolor{red}{\underline{action}} \textcolor{red}{\underline{party}} .}\\\\

    \end{tabularx}
    }
    \\
    \subfloat[][\rmfamily\label{tab:scoring_quali2}]{
    \ttfamily
	\begin{tabularx}{\linewidth}{lX}

		\toprule
		\textsc{key}   & \textsc{value} \\
		\midrule
		name           & ryan moore \\
		spouse         & nichole olson -lrb- m. 2011 -rrb- \\
		children       & tucker \\
		college        & unlv \\
		yearpro        & 2005 \\
		tour           & pga tour \\
		prowins        & 4 \\
		pgawins        & 4 \\
		masters        & t12 2015 \\
		usopen         & t10 2009 \\
		open           & t10 2009 \\
		pga            & t9 2006 \\
		article\_title & ryan moore -lrb- golfer -rrb- \\
		\bottomrule
		\\
		\multicolumn{2}{p{.9\linewidth}}{\textbf{Ref.}: ryan david moore -lrb- born december 5 , 1982 -rrb- is an american professional golfer , currently playing on the pga tour .}\\
		\multicolumn{2}{p{.9\linewidth}}{\textbf{PB\&L}: ryan \textcolor{red}{\underline{david}} moore \textcolor{red}{\underline{-lrb-}} \textcolor{red}{\underline{born}} \textcolor{red}{\underline{december}} \textcolor{red}{\underline{5}} \textcolor{red}{\underline{,}} \textcolor{red}{\underline{1982}} \textcolor{red}{\underline{-rrb-}} \textcolor{red}{\underline{is}} \textcolor{red}{\underline{an}} \textcolor{red}{\underline{american}} \textcolor{red}{\underline{professional}} \textcolor{red}{\underline{golfer}} \textcolor{red}{\underline{,}} \textcolor{red}{\underline{currently}} \textcolor{red}{\underline{playing}} \textcolor{red}{\underline{on}} \textcolor{red}{\underline{the}} \textcolor{red}{\underline{pga}} \textcolor{red}{\underline{tour}} \textcolor{red}{\underline{.}}}\\
		\multicolumn{2}{p{.9\linewidth}}{\textbf{Ours}: ryan david moore \textcolor{red}{\underline{-lrb-}} \textcolor{red}{\underline{born}} \textcolor{red}{\underline{december}} \textcolor{red}{\underline{5}} \textcolor{red}{\underline{,}} \textcolor{red}{\underline{1982}} \textcolor{red}{\underline{-rrb-}} is an american professional golfer , currently playing on the pga tour .}\\\\

	\end{tabularx}
    }
	\caption{WikiBio instances' hallucinated words according either to our scoring procedure or to the method proposed by \citet{PerezBeltrachini2018bootstrap}. \textit{PB\&L} labels word incoherently (a), and sometimes the whole reference text (b). In comparison, our approach leads to a fluent breakdown of the sentences in hallucinated/factual statements.}
	\label{tab:scoring_quali}

\end{figure*}

\vspace{2mm}
\noindent\textbf{Intrinsic performance.}
Tab.~\ref{tab:scorecompare} (top) compares the labeling performance of our method and  \textit{PB\&L} against human judgment.
Our scoring procedure significantly improves over \textit{PB\&L}: the latter only achieves $46.9$\% accuracy and $29.7$\% F-measure, against $87.5$\% and $68.7$\% respectively for our proposed procedure. \citet{PerezBeltrachini2018bootstrap} report a F-measure of $36$\%, a discrepancy that can be explained by the difference between the evaluation procedures: \textit{PB\&L} evaluate on 132 sentences, several of which can be tied to the same table, whereas we explicitly chose to evaluate on $300$ sentences all from different tables in order to minimize correlation.

We remark that beyond F-measure, the precision of \textit{PB\&L}'s scoring procedure is at $21.3$\% compared to $80.6$\% for ours, and recall stands at $49.2$\% against $59.8$\%.
We argue that selecting a negative instance at random for training their classifier leads the network to incoherently label words, without apparent justification. See Figure~\ref{tab:scoring_quali} for two examples of this phenomenon; and Appendix~\ref{appendix-sec:model-outputs} for other comparisons.
In contrast, our method is able to detect hallucinated statements inside a sentence, without incorrectly labeling the whole sentence as hallucinated.

\vspace{2mm}
\noindent\textbf{Impact on a DTG downstream task.}
Additionally, we assess the difference of both scoring procedures using their impact on the WikiBio DTG task. Specifically, Tab.~\ref{tab:scorecompare} (bottom) shows the results of training \textit{MBD} using either \textit{PB\&L}'s or our labels. We observe significant improvements,
especially in BLEU and PARENT-recall ($40.5$\% vs $32.2$\% and $45$\% vs $39.3$\%), showing that our labeling procedure is more helpful at retaining information from training instances (the system better picks up what humans picked-up, ultimately resulting in better BLEU and recall). 
\begin{table*}

    \scriptsize
    \begin{tabularx}{\textwidth}{lccccYYc}
    \toprule
    \multirow{3}*{Model} & \multirow{3}*{BLEU\textsuperscript{$\uparrow$}} & \multicolumn{3}{c}{PARENT\textsuperscript{$\uparrow$}} & \multirow{3}={\centering Halluc. rate\textsuperscript{$\downarrow$}} & \multirow{3}={\centering Mean sent. length} & \multirow{3}*{Flesch\textsuperscript{$\downarrow$}} \\
    \cmidrule(lr){3-5}
    && Precision & Recall & F-measure \\
    \midrule
    Gold                    & -       & -       & -       & -       & 23.82\% & 19.20 & \bf 53.80\% \\
    \texttt{stnd}           & 41.77\% & 79.75\% & 45.02\% & 55.28\% &  4.20\% & 13.80 & 58.90\% \\
    \texttt{stnd\_filtered} & 34.66\% & \bf80.90\% & 42.48\% & 53.27\% & \bf0.74\% & 12.00 & 62.10\% \\
    \texttt{hsmm}          & 35.17\% & 71.72\% & 39.84\% & 48.32\% & 7.98\% & 14.80 & 58.60\% \\
    \texttt{hier}           & \bf45.14\% & 75.09\% & 46.02\% & 54.65\% & 10.10\% & 16.80 & 56.20\% \\
    \texttt{hal\textsubscript{WO}} & 36.50\% & 79.50\% & 40.50\% & 51.70\% & - & - & - \\
    \midrule
    \texttt{MBD} & 41.56\% & 79.00\% & \bf46.40\% & \bf56.16\% & 1.43\% & 14.60 & 58.80\% \\
    \bottomrule
    \end{tabularx}
    \caption{Comparison results on WikiBio.  \textsuperscript{ $\uparrow$ } (resp. \textsuperscript{ $\downarrow$ }) means higher (resp. lower) is better. ``Gold'' refers to the gold standard, i.e.~the reference texts included in the dataset.}
    \label{tab:auto-eval}
        \vspace{-0.5cm}
\end{table*}

    \vspace{-0.5cm}
\subsection{Automatic System Evaluation}
\label{subsect:auto-eval}
    \vspace{-0.2cm}

\noindent\textbf{Comparison with SOTA systems.}
Tab.~\ref{tab:auto-eval} shows the performances of our model and all baselines
according to the metrics of
Sec.~\ref{subsec:metrics}. Two qualitative examples are presented in Figure~\ref{fig:ex_quali_wikibio} and more are available in Appendix~\ref{appendix-sec:model-outputs}.

First of all,
reducing hallucinations is reached with success, as highlighted by the hallucination rate ($1.43\%$ vs.  $4.20\%$ for a standard encoder-decoder and $10.10\%$ for the best SOTA model on BLEU).
The only model which gets a lower hallucination rate ($0.74\%$, corroborated by its PARENT-precision of $80.9\%$), \textit{stnd\_filtered}, achieves such a result at a high cost. As can be seen in Figure~\ref{fig:ex_quali_wikibio} where its output is factual but cut short, its sentences are the shortest and the most naive in terms of the Flesch readability index, which is also reflected by a lower BLEU score.
The high PARENT precision -- mostly due to the shortness of the outputs -- is counterbalanced by a low recall: the F-measure indicates the overall lack of competitiveness of this trade-off. This shows that the naive approach of simply filtering training instances is not the appropriate solution for hallucination reduction. This echoes \citep{Filippova2020hallucinations} who trained a vanilla network on the cleanest $20$\% of the data and found that predictions are more precise than those of a model trained on $100$\% but that PARENT-recall and BLEU scores are low.

At the other extreme, the best model in terms of BLEU, \textit{hier}, falls short regarding precision, suggesting that often the generated text is not matched in the input table; this issue is also reflected by the highest hallucination rate of all models ($10.10\%$). A reason could be the introduction of their auxiliary training tasks which often drive the decoder to excess in mimicking human behavior. While BLEU score improves, overall factualness of outputs decreases, showing that the model picks up domain lingo (how to formulate ideas) but not domain insight (which ideas to formulate) (see Figure~\ref{fig:ex_quali_wikibio}).
This is in line with \citep{Reiter2018,Filippova2020hallucinations} who argue that BLEU is an inappropriate metric for generation tasks other than machine translation.

The analysis of \textit{hsmm}, and especially of its relatively weak performance both in terms of BLEU and PARENT, highlights the insufficiency of the multi-branch architecture by itself. This reinforces the need of the additional hallucinations supervision provided by our labeling procedure.

\begin{figure*}
\scriptsize
\subfloat[][\label{fig:ex_quali_wikibio1}]{\parbox{\textwidth}{
\centering
\begin{tabular}{ll}
\toprule
    \texttt{name} & \texttt{zack lee} \\
    \texttt{birth\_name} & \texttt{zack lee jowono} \\
    \texttt{nationality} & \texttt{indonesian} \\
    \texttt{occupation} & \texttt{actor , boxer , model} \\
    \texttt{birth\_date} & \texttt{15 august 1984} \\
    \texttt{birth\_place} & \texttt{liverpool , merseyside , england , uk} \\
    \texttt{years\_active} & \texttt{2003 -- present} \\
    \texttt{parents} & \texttt{hendra and ayu jowono} \\
    \texttt{spouse} & \texttt{nafa urbach ( 2007 -- present )} \\
    \texttt{article\_title} & \texttt{zack lee} \\
\end{tabular}
\\
\begin{tabularx}{\textwidth}{lX}
\toprule
    Gold          & zack lee ( born 15 august 1984 ) is an indonesian actor , model and boxer of british descent . \\
    \texttt{stnd} & zack lee jowono ( born 15 august 1984 ) is an indonesian actor and model . \\
    \texttt{stnd\_filtered} & zack lee ( born zack lee jowono ; 15 august 1984 ) is an indonesian actor . \\
    \texttt{hsmm} & zack lee jowono ( born 15 august 1984 ) is an indonesian actor who has appeared in tamil films . \\
    \texttt{hier} & zack lee jowono ( born 15 august 1984 ) , better known by his stage name zack lee , is an indonesian actor , model and model . \\
    \texttt{MBD[.4, .1, .5]} & zack lee ( born zack lee jowono ; 15 august 1984 ) is an indonesian actor , boxer and model . \\
\bottomrule
\end{tabularx}
}}
\\
\subfloat[][\label{fig:ex_quali_wikibio2}]{\parbox{\textwidth}{
\centering
\begin{tabular}{ll}
\toprule
    \texttt{name} & \texttt{wayne r. dynes} \\
    \texttt{birth\_date} & \texttt{23 august 1934} \\
    \texttt{occupation} & \texttt{professor , historian , and encyclopedist} \\
    \texttt{article\_title} & \texttt{wayne r. dynes} \\
\end{tabular}

\begin{tabularx}{\textwidth}{lX}
\toprule
    Gold          & wayne r. dynes ( born august 23 , 1934 ) is an american art historian , encyclopedist , and bibliographer . \\
    \texttt{stnd} & wayne r. dynes ( born august 23 , 1934 ) is an american historian and encyclopedist . \\
    \texttt{stnd\_filtered} & wayne r. dynes is a professor . \\
    \texttt{hsmm} & wayne r. dynes ( born august 23 , 1934 ) is an american historian , historian and encyclopedist . \\
    \texttt{hier} & wayne r. dynes ( born august 23 , 1934 ) is an american professor of history at the university of texas at austin . \\
    \texttt{MBD[.4, .1, .5]} & wayne r. dynes ( born august 23 , 1934 ) is an american professor , historian , and encyclopedist . \\
\bottomrule
\end{tabularx}
}}
\caption{Qualitative examples of our model and baselines on the WikiBio test set. Note that: (1) \textit{gold} references may contain divergences; (2) \textit{stnd} and \textit{hsmm} seem to perform well superficially, but often hallucinate; (3) \textit{stnd\_filtered} doesn't hallucinate but struggles with fluency;  (4) \textit{hier} overgenerate "human-sounding" statements, that lacks facutalness;     (5) \textit{MBD} sticks to the fact contained by the table, in concise and fluent sentences.       }
\label{fig:ex_quali_wikibio}
    \vspace{-0.4cm}
\end{figure*}

Finally, in the comparisons with
\textit{hal\textsubscript{WO}}, we can see that while it
achieves one of the highest performances in term of precision ($79.5$\%), this comes at the cost of the lowest recall ($40.5$\%) of all models and thus poor F-measure. This confirms our hypothesis that, while effective at producing mostly factual content, modeling hallucination only as a fixed value
for a whole instance is detrimental to the content generation procedure. Finer-grain annotations are required, as shown by our model recall ($46.4$\%), coupled with a robust  precision ($79.0$\%).

\vspace{2mm}
\noindent\textbf{Weight impact on decoding.}
\label{subsec:weights-impact}
As we deal with a CTG system, we can guide our network
at inference to generate sentences following desired attributes. The impact of different weight combinations is explored in Tab.~\ref{tab:different-weight-combinations}. In particular,
we can see that changing weights in favor of the hallucination factor (top five lines)
leads to decreases in both precision and recall (from $80.37$\% to $57.88$\% and $44.96$\% $4.82$\% respectively).
We also observe that strongly relying on the hallucinating branch dramatically impacts performances ([$0.0$~$0.5$~$0.5$] obtains near 0 BLEU and F-measure), as it is never fed with complete, coherent sentences during training. However, some
performance can still be restored via the fluency branch: [$0.0$~$0.1$~$0.9$] performs at $15.51$\% BLEU and $36.88$\% F-measure.

\begin{table}
    \centering
    \begin{tabular}{ccccc}
    \toprule
    \multirow{3}*{Weights} & \multirow{3}*{BLEU\textsuperscript{$\uparrow$}} & \multicolumn{3}{c}{PARENT\textsuperscript{$\uparrow$}}\\
    \cmidrule(lr){3-5}
    && Precision & Recall & F-measure \\
    \midrule
    $0.5$~~$0.0$~~$0.5$ & 38.90\%  & 80.37\% & 44.96\% & 55.29\%  \\
    $0.4$~~$0.1$~~$0.5$ & 41.56\%  & 79.00\% & 46.40\% & 56.16\%  \\
    $0.3$~~$0.2$~~$0.5$ & 42.68\%  & 72.99\% & 45.81\% & 53.74\%  \\
    $0.2$~~$0.3$~~$0.5$ & 22.64\%  & 53.92\% & 32.96\% & 36.55\%  \\
    $0.1$~~$0.4$~~$0.5$ & 2.03\% & 57.88\% & 4.82\%  & 6.79\%  \\
    $0.0$~~$0.5$~~$0.5$ & 0.32\% & 85.01\% & 1.02\%  & 1.78\%  \\
    $0.0$~~$0.4$~~$0.6$ & 1.07\% & 62.71\% & 2.47\%  & 3.66\%  \\
    $0.0$~~$0.3$~~$0.7$ & 2.81\% & 42.86\% & 6.15\%  & 7.94\%  \\
    $0.0$~~$0.2$~~$0.8$ & 7.30\%  & 41.78\% & 16.58\% & 18.68\%  \\
    $0.0$~~$0.1$~~$0.9$ & 15.51\%  & 56.93\% & 32.85\% & 36.88\%  \\
    \bottomrule
    \end{tabular}

    \caption{Performances of \textit{MBD} on WikiBio validation set, with various weight settings. Weights' order is (\textit{content}, \textit{hallucination}, \textit{fluency}).}
    \label{tab:different-weight-combinations}
        \vspace{-0.4cm}
\end{table}

It is interesting to note that the relaxation of the strict constraint on the content factor in favor of the hallucination factor, ([$0.4$~$0.1$~$0.5$] $\rightarrow $ [$0.5$~$0.0$~$0.5$]) obtains better performances ($56.16$\% vs $55.29$\% F-measure). This highlights that strictly constraining on content yields sensibly more factual outputs ($79$\% vs $80.37$\% precision), at the cost of constraining the model's generation creativity ($46.40$\% vs $44.96$\% recall). The [$0.4$~$0.1$~$0.5$] variant has more ``freedom of speech'' and sticks more faithfully to domain lingo (recall and BLEU), without compromising too much in terms of content.

    \vspace{-0.4cm}
\subsection{Human evaluation}
\label{subsec:human-evaluation}
    \vspace{-0.2cm}

\begin{table}
    \centering
    \begin{tabular}{lccc}
    \toprule
    Model                  & Fluency   & Factualness & Coverage \\
    \midrule
    Gold                    &    98.7\% & 32.0\%    & \bf 4.47 \\
    \texttt{stnd\_filtered} &    93.5\% & \bf86.1\% &     4.07 \\
    \texttt{hier}           &    97.4\% & 55.0\%    &     4.45 \\
    \texttt{MBD}            & \bf99.6\% & 76.6\%    &     4.46 \\
    \bottomrule
    \end{tabular}

    \caption{Results of the human evaluation on WikiBio\protect\footnotemark.}
    \label{tab:human-evaluation}
        \vspace{-0.6cm}
\end{table}
\footnotetext{Fluency reports the sum of ``fluent'' and ``mostly fluent'',  as ``mostly fluent'' often comes from misplaced punctuation and doesn't really impact readability. However, Factualness reports only the count of ``factual'', as ``mostly factual'' sentences contain hallucinations and cannot be considered ``factual''.}

To measure subtleties which are not captured by automatic metrics, we report in Tab.~\ref{tab:human-evaluation} human ratings of our model, two baselines and the gold. These baselines have been selected because they showcase interesting behaviors on automatic metrics:
\textit{hier} obtains the best BLEU score but a poor precision, and \textit{stnd\_filtered} gets the best precision but poor BLEU, length and Flesch index.

First, coherently with \citep{Dhingra2019},
we found that around two thirds of gold references contain divergences from their associated tables. Such data also confirm our analysis on the \textit{stnd\_filtered} baseline: it's training on truncated sentences lead to an unquestionable ability to avoid hallucinations, while dramatically impacting both its fluency and coverage, leading to less desired outputs overall, despite the high PARENT-precision score.\\

The comparison between \textit{hier} and \textit{MBD} shows that both approaches lead to similar coverage, with \textit{MBD} obtaining significantly better performances in terms of factualness. We also highlight that \textit{MBD} is evaluated as being the most fluent one, even better than the reference (which can be explained by the imperfect pre-processing done by \citet{Lebret2016}).

\begin{table*}
    \tiny
    \begin{tabularx}{\textwidth}{Xccccccc}
    \toprule
    \multirow{3}*{Model} & \multirow{3}*{BLEU\textsuperscript{$\uparrow$}} & \multicolumn{3}{c}{PARENT\textsuperscript{$\uparrow$}} & \multicolumn{3}{c}{Human evaluation} \\
    \cmidrule(lr){3-5} \cmidrule(lr){6-8}
    && Precision & Recall & F-measure & Fluency\textsuperscript{$\uparrow$} & Factualness\textsuperscript{$\uparrow$} & Coverage \\
    \midrule
    Gold(noisy)                   & - & - & - & -                        & 97.1\% (97.1) & 91.2\% (79.4) & 3.618 \\
    \texttt{stnd}                  & \bf21.27\% & 56.60\% & 25.16\% & \bf29.71\% & 55.9\% (26.5) & 53.0\% (20.6) & 2.824 \\
    \texttt{stnd\_filtered}        & 19.48\% & 56.69\% & 22.31\% & 27.18\% & 29.4\% (8.8) & 70.6\% (50.0) & 2.706 \\
    \texttt{hal\textsubscript{WO}} & 17.06\% & \bf77.64\% & 22.65\% & 29.38\% & 61.7\% (38.2) & 61.8\% (32.4) & 2.725 \\
    \midrule
    \texttt{MBD}                   & 18.35\% & 50.44\% & \bf25.25\% & 28.25\% & \bf91.2\% (50.0) & \bf85.3\% (55.9) & \bf3.613 \\
    \bottomrule
    \end{tabularx}

    \caption{Comparison results on ToTTo.  \textsuperscript{ $\uparrow$ } (resp. \textsuperscript{ $\downarrow$ }) means higher (resp. lower) is better. In human evaluation for Fluency, reported are for ``Fluent'' and ``Mostly Fluent'', with only ``Fluent'' in parentheses. Same for Factualness.}
    \label{tab:totto-auto-eval}

\end{table*}

\vspace{-0.4cm}
\subsection{ToTTo: a considerably noisy setting}
\vspace{-0.2cm}

\begin{figure*}
\scriptsize
\subfloat[][\label{fig:totto-quali-ex_1}]{\parbox{\textwidth}{
\centering
\begin{tabular}{ll}
\toprule
    \texttt{page\_title} & \texttt{Huge (TV series)} \\
    \texttt{section\_title} & \texttt{Episodes} \\
    \texttt{Original\_air\_date} & \texttt{June 28 2010} \\
    \texttt{U.S.\_viewers\_(millions)} & \texttt{2.53} \\
\end{tabular}
\\
\begin{tabularx}{\textwidth}{lX}
\toprule
    Gold (clean)       & The TV series , Huge , premiered on June 28 , 2010 with 2.53 million viewers. \\
    Gold (noisy)       & The series premiered on June 28 , 2010 at 9 p.m. with 2.53 million viewers . \\
    \texttt{stnd} & On June 28 , 2010 , it was watched by 2.53 million viewers . \\
    \texttt{stnd\_filtered} & was watched by 2.53 on June 28 , 2010 . \\
    \texttt{hal\textsubscript{WO}} & June 28 , 2010 : Huge million viewers . \\
    \texttt{MBD[.4, .1, .5]} & Huge 's first episode , aired on June 28 , 2010 , was watched by 2.53 million . \\
\bottomrule
\end{tabularx}
}}

\subfloat[][\label{fig:totto-quali-ex_2}]{\parbox{\textwidth}{
\centering
\begin{tabular}{ll}
\toprule
    \texttt{page\_title} & \texttt{LM317} \\
    \texttt{section\_title} & \texttt{Specification} \\
    \texttt{Parameter} & \texttt{Output voltage range} \\
    \texttt{Value} & \texttt{1.25 - 37} \\
\end{tabular}
\\
\begin{tabularx}{\textwidth}{lX}
\toprule
    Gold (clean)         & LM317 produces a voltage of 1.25 V . \\
    Gold (noisy)         & Internally the device has a bandgap voltage reference which produces a stable reference voltage of Vref= 1.25 V followed by a feedback-stabilized amplifier with a relatively high output current capacity . \\
    \texttt{stnd} & The Output is a Output range of 1.25 – 37 . \\
    \texttt{stnd\_filtered} & range from 1.25 to 37 . \\
    \texttt{hal\textsubscript{WO}} & Output voltage range 1.25 – 37 – 37 . \\
    \texttt{MBD[.4, .1, .5]} & The Output 's range is approximately 1.25 . \\
\bottomrule
\end{tabularx}
}}
\vspace{-0.2cm}
\caption{Qualitative examples of \textit{MBD} and \textit{hal\textsubscript{WO}} on ToTTo. \textit{hal\textsubscript{WO}}'s poor generation quality is not detected by discrete metrics. In contrast, \textit{MBD} generates fluent and naively factual sentences. Note that \textit{stnd} and \textit{stnd\_filtered} have the same behavior as on WikiBio: the former produces fluent but nonsensical text; the latter generates very un-fluent, but factual, text.}
\label{fig:totto-quali-ex}
    \vspace{-0.4cm}
\end{figure*}

The ToTTo dataset is used in the following experiments to explore models' robustness to the impact of extreme noise during training. As stated in Section~\ref{subsec:datasets}, we use as inputs only the \textit{highlighted} cells, as content selection is arbitrary (i.e.~the cells were chosen depending on the target sentence, and not vice versa). On the other hand, we use as targets the noisy references, which may contain both divergences and lexical issues. This setting is particularly challenging and is more effective in recreating a representational, hallucination-prone real-life context than WikiBio.
Other datasets~\citep{Novikova2017e2e,Gardent2017,Wen2015} available in literature are too similar to WikiBio concerning their goals and challenges, and are therefore less interesting in this context.

Table~\ref{tab:totto-auto-eval} reports the performances of \textit{stnd}, \textit{stnd\_filtered}, \textit{hal\textsubscript{WO}} and \textit{MBD} with regards to automatic metrics and human evaluation. Compared to their respective performances on WikiBio, all models show significantly decreased scores. They struggle at generating syntactically correct sentences but, at the same time, they have still learned to leverage their copy mechanism and to stick to the input. This behavior is illustrated in both examples of Fig.~\ref{fig:totto-quali-ex}. In particular, \textit{hal\textsubscript{WO}}'s high PARENT-precision score ($77.64$\%) seems to be due to its tendency to blindly copy input data without framing them in a sentence structure, as its low BLEU and PARENT-recall scores suggests ($17.06$\% and $22.65$\%).
These lower scores are good indicators that the ToTTo task, as framed in this paper, is difficult. Following the same evaluation protocol than for WikiBio, we report human ratings of different models, also included in Table~\ref{tab:totto-auto-eval}.

\textit{MBD}'s factualness is judged favorably, with $55.9$\% hallucination-free texts, and up to $85.3$\% texts with a single error at most.
In contrast, \textit{hal\textsubscript{WO}} stands at $32.4$\% and $61.8$\% for error-free texts and single-error texts respectively. Interestingly, \textit{stnd\_filtered} obtains the second best performance ($70.6$\% texts with a single error).

Fluency scores are also meaningful: \textit{hal\textsubscript{WO}} and \textit{MBD} respectively obtain $61.7$\% and $91.2$\%.
Word-based filtering is not suitable for noisy datasets, as shown by \textit{stnd\_filtered}'s worse fluency score, $29.4$\%.

As for coverage performances, our model \textit{MBD} obtains the maximum coverage score $3.613$, surpassing all baselines by at least $0.789$ slots (the second best coverage score is obtained by \textit{stnd} at $2.824$), and getting very close to the Gold value (which stands at $3.618$). These performances, and qualitative examples of Figure~\ref{fig:totto-quali-ex}, suggest that \textit{stnd\_filtered} and \textit{hal\textsubscript{WO}} try to reduce hallucinations at the cost of missing some input slot, while \textit{MBD} effectively balances both goals.

The analysis of Factualness, Fluency and Coverage can be enhanced using qualitative error analysis on randomly sampled generated texts (we report two such examples in Figure~\ref{fig:totto-quali-ex}). In particular, we want to highlight the following considerations:

\begin{itemize}
    \item
    As most training examples are very noisy, sentence-level models fail at learning from them.
    \textit{stnd\_filtered} has been trained on factual statements only, at the cost of using mostly incomplete sentences during training. On both examples of Figure~\ref{fig:totto-quali-ex}, it generated truncated sentences, missing their subjects. Its relatively high Factualness and low Fluency scores indicate that it did not learn to produce diverging outputs, nor complete sentences.
    Differently, \textit{hal\textsubscript{WO}} generates incorrectly ordered sequences of words extracted from the table (Fig.~\ref{fig:totto-quali-ex_1}), or repetitions (Fig.~\ref{fig:totto-quali-ex_2}).
    The low number of training instances containing the input pair \textit{(hallucination ratio, 0)} does not allow to learn what a non-hallucinated sentence actually consists in.
    \item In contrast, our proposed finer-grained approach proves helpful in this setting, as shown by the human evaluation: sentences generated by \textit{MBD} are more fluent and more factual. The multi-branch design enables the model to leverage the most of each training instance, leading to better performances overall.
    \item Finally, we acknowledge that despite over-performing other models, \textit{MBD} obtains only $55.9$\% of \textit{factual} sentences. For instance, in Figure~\ref{fig:totto-quali-ex_2}, our model does not understand that a range consists of two numbers. The difficulty of current models to learn on very noisy and diverse datasets shows that there is still room for improvement in hallucination reduction in DTG.
\end{itemize}

\vspace{-0.4cm}
\section{Conclusion}
\label{sec:conclusion}

We proposed a Multi-Branch decoder, able to leverage word-level alignment labels in order to produce factual and coherent outputs. Our proposed labeling procedure is more accurate than previous work, and outputs from our model are estimated, by automatic metrics and human judgment alike, more fluent, factual, and relevant. We obtain state-of-the-art performances on WikiBio for PARENT F-measure, and show that our approach is promising in the context of a noisier setting.

We designed our alignment procedure to be general and easily reproducible on any DTG dataset. One strength of our approach is that co-occurrences and dependency parsing can be used intuitively to extract more information from the tables than a naive word matching procedure.
However, in the context of tables mainly including numbers (e.g., RotoWire), the effectiveness of the co-occurrence analysis is not guaranteed.
A future work will be to improve upon the co-occurrence analysis to generalize to tables which contain less semantic inputs. For instance, the labeling procedure of \citet{PerezBeltrachini2018bootstrap} might be revised  so that adverse instances are not selected randomly, which we hypothesize would result in more relevant labels.

Finally, experiments on ToTTo outline the  narrow exposure to language of current models when used on very noisy datasets. Our model has shown interesting properties through the human evaluation but is still perfectible.
Recently introduced large pretrained language models, which have seen significantly more varied texts, may attenuate this problem. In this direction, adapting the work of \citep{chen2020-fewshot,kale2020-pretrain} to our model could bring improvements to the results presented in this paper.

\vspace{-0.6cm}
\begin{small}
\section*{Declarations}
\vspace{-0.4cm}
$\bullet$ Fundings: We would like to thank the H2020 project AI4EU (825619) and the ANR JCJC SESAMS (ANR-18-CE23-0001) for supporting this work.
This research has been partially carried on in the context of the Visiting Professor Program of the Italian Istituto Nazionale di Alta Matematica (INdAM).\\
$\bullet$ Conflict of interest: No conflict of interest\\
$\bullet$ Code availability: Code is available at \url{https://github.com/KaijuML/dtt-multi-branch}\\
$\bullet$ Other items are not applicable (Availability of data and material, Additional declarations for articles in life science journals that report the results of studies involving humans and/or animals, Ethics approval, Consent to participate, Consent for publication)
\end{small}

\vspace{-0.4cm}
\bibliographystyle{spmpscinat.bst}
\bibliography{refs_no_url.bib}

\clearpage
\appendix

\section{Alignment labels reproducibility}
\label{appendix-sec:alignment-labels}

We consider as \textit{important words}, i.e.~nouns, adjectives or numbers, those which are Part-of-Speech tagged as \texttt{NUM}, \texttt{ADJ}, \texttt{NOUN} and \texttt{PROPN}.

In order to apply the score normalization function $\mathit{norm(\cdot, y)}$, we separate sentences $y$ into statements $y_{t_i:t_{i+1}-1}$. To do so, we identify the set of introductory dependency relation labels\footnote{\texttt{acl}, \texttt{advcl}, \texttt{amod}, \texttt{appos}, \texttt{ccomp}, \texttt{conj}, \texttt{csubj}, \texttt{iobj}, \texttt{list}, \texttt{nmod}, \texttt{nsubj}, \texttt{obj}, \texttt{orphan}, \texttt{parataxis}, \texttt{reparandum}, \texttt{vocative}, \texttt{xcomp}; every dependency relation is documented in the \href{https://universaldependencies.org/u/dep/index.html}{Universal Dependencies website}.}, following previous work on rule-based systems for the conversion of dependency relations trees to constituency trees \citep{Han2000, Xia2001, Hwa2005, Borensztajn2009}.

Our segmentation algorithm considers every leaf token in the dependency tree, and seeks its nearest ancestor which is the root of a statement.

Two heuristics enforce the score normalization:
\begin{enumerate*}[label=(\roman*)]
    \item conjunctions and commas next to hallucinated tokens acquires these lasts' hallucination scores, and
    \item paired parentheses and quotes acquire the minimum inner tokens' hallucination score.
\end{enumerate*}

Part-of-Speech tagging has been done using the HuggingFace's Transformers library \citep{huggingface} to fine-tune a BERT model \citep{Devlin2019} on the UD English ParTUT dataset \citep{Sanguinetti2015}; Stanza \citep{stanza} has been exploited for dependency parsing.

\section{Implementation details}
\label{appendix-sec:implementation-details}
Our system is implemented in Python 3.8\footnote{\url{http://www.python.org}} and PyTorch 1.4.0\footnote{\url{http://www.pytorch.org}}. In particular, our multi-branch architecture is developed, trained and tested as an OpenNMT \citep{opennmt} model. Sentence lengths and Flesch index \citep{flesch} are computed using the standard \texttt{style} Unix command.

Differently to \citet{PerezBeltrachini2018bootstrap}, we did not adapt the original WikiBio dataset\footnote{\url{https://github.com/DavidGrangier/wikipedia-biography-dataset}} in any manner: as we work on the model side, we fairly preserve the dataset's noisiness.

Word-level alignment labels are computed setting $m=5$, following \citet{Mikolov2013}. As stated in Sec.~\ref{subsec:implementation-details}, the threshold $\tau$'s value is optimized for highest accuracy via human tuning: Table~\ref{tab:human-tune} shows accuracy scores of our proposed automated labels for different values of $\tau$.

We share the vocabulary between input and output, limiting its size to 20000 tokens. Hyperparameters were tuned using performances on the development set:  Tab.~\ref{tab:validation} reports the performances of our best performing \textit{MBD} on the development set. Our encoder consist of a 600-dimensional embedding layer followed by a 2-layered bidirectional LSTM network with hidden states sized 600. We use the \textit{general} attention mechanism with input feeding \citep{Luong2015} and the same copy mechanism as \citet{See2017}. Each branch of the multi-branch decoder is a 2-layered LSTM network with hidden states sized 600 as well.

Training is performed using the Adam algorithm \citep{Kingma2015} with learning rate $\eta=10^{-3}$, $\beta_1=0.9$ and $\beta_2=0.999$. The learning rate is decayed with a factor of $0.5$ every 10000 steps, starting from the 5000th one. We used minibatches of size $64$ and regularized via clipping the gradient norm to 5 and using a dropout rate of $0.3$. We used beam search during inference, with a beam size of 10.

All experiments were performed on a single NVIDIA Titan XP GPU. Number of parameters and training times are shown in Table~\ref{tab:sizetime}. Same model's differences between WikiBio and ToTTo are justified by the different datasets' number of instances and input vocabulary sizes.

\section{Annotation interface}
\label{sec:interface}
The human annotation procedure is done via a web application specifically developed for this research. Fig.~\ref{fig:tagging} shows how the hallucination tagging user interface looked like in practice, while in Fig.~\ref{fig:ranking} a typical sentence analysis page is shown.

\setcounter{table}{0}
\renewcommand{\thetable}{B\arabic{table}}
\begin{table*}
    \centering
    \begin{tabular}{ccccc}
    \toprule
    Threshold & Accuracy & F-measure & Precision & Recall \\
    \midrule
    0.0 & 70.2\% & 56.8\% & 42.2\% & 86.7\% \\
    0.4 & \bf86.0\% & \bf70.6\% & 67.0\% & 74.6\% \\
    0.8 & 85.8\% & 62.8\% & 77.3\% & 52.9\% \\
    \bottomrule
    \end{tabular}
    \caption{Accuracy scores of our proposed word-level automated labels for different values of the threshold $\tau$.}
    \label{tab:human-tune}
\end{table*}
\begin{table*}
    \centering
    \begin{tabular}{lcccc}
    \toprule
    \multirow{3}*{Model} & \multirow{3}*{BLEU} & \multicolumn{3}{c}{PARENT}\\
    \cmidrule(lr){3-5}
    && Precision & Recall & F-measure \\
    \midrule
    \texttt{MBD[.4, .1, .5]} & 42.50\% & 79.26\% & 46.09\% & 55.95\% \\
    \bottomrule
    \end{tabular}
    \caption{The performances of our model on the WikiBio validation set.}
    \label{tab:validation}
\end{table*}
\begin{table}
    \centering
    \begin{tabular}{llcc}
    \toprule
    Dataset & Model                   & Size [M] & Training time [h] \\
    \midrule
    \multirow{4}*{WikiBio}  & \texttt{stnd}                     & $41$ & ~$5$ \\
                            & \texttt{stnd\_filtered}           & $41$ & ~$5$ \\
                            & \texttt{hal\textsubscript{WO}}    & $41$ & ~$5$ \\
                            & \texttt{MBD}                      & $55$ & $10$ \\
    \midrule
    \multirow{4}*{ToTTo}    & \texttt{stnd}                     & $62$ & ~$4$ \\
                            & \texttt{stnd\_filtered}           & $62$ & ~$4$ \\
                            & \texttt{hal\textsubscript{WO}}    & $62$ & ~$4$ \\
                            & \texttt{MBD}                      & $76$ & ~$8$ \\
    \bottomrule
    \end{tabular}

    \caption{Sizes and training times of the implemented models.}
    \label{tab:sizetime}
\end{table}

\setcounter{figure}{0}
\renewcommand{\thefigure}{C\arabic{figure}}

\begin{figure*}
    \centering
    \subfloat[][Hallucination tagging]
    {\label{fig:tagging}\includegraphics[width=\textwidth]{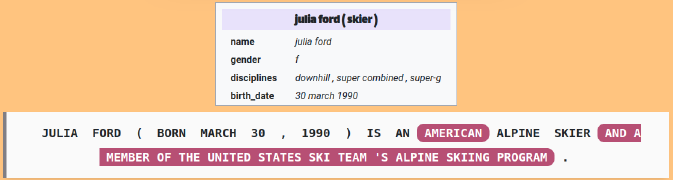}}\\
    \subfloat[][Sentence analysis]
    {\label{fig:ranking}\includegraphics[width=\textwidth]{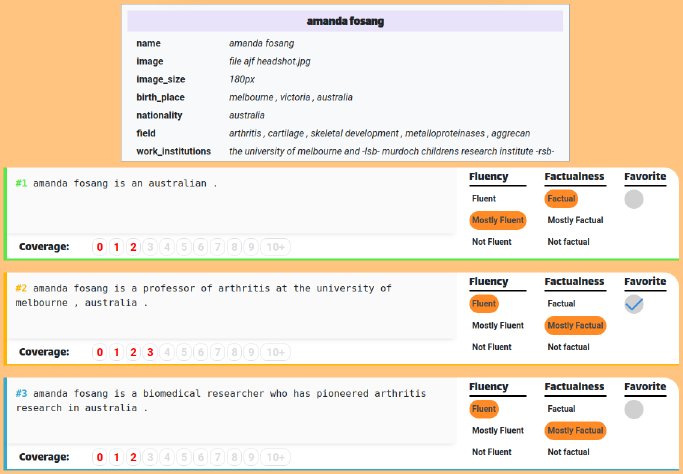}}
    \caption{The human annotation tasks, as presented to the annotators.}
    \label{fig:human_gui}
\end{figure*}

\section{Qualitative examples}
\label{appendix-sec:model-outputs}

Tables~\ref{tab:ex_scoring1} and \ref{tab:ex_scoring2} show word-level labeling of WikiBio training examples. Underlined, red words are hallucinated according either to our scoring procedure or to the method proposed by \citet{PerezBeltrachini2018bootstrap}.

\setcounter{table}{0}
\renewcommand{\thetable}{D\arabic{table}}

\begin{table*}
	\centering
	\ttfamily
	\normalsize
	\begin{tabularx}{\linewidth}{lX}
		\toprule
		\textsc{key}   & \textsc{value} \\
		\midrule
		name           & susan blu \\
		birth\_name    & susan maria blupka \\
		birth\_date    & 12 july 1948 \\
		birth\_place   & st paul , minnesota , u.s.            \\
		occupation     & actress , director , casting director \\
		yearsactive    & 1968 -- present \\
		article\_title & susan blu \\
		\bottomrule
		\\
		\multicolumn{2}{p{.93\linewidth}}{\textbf{Ref.}: susan maria blu -lrb- born july 12 , 1948 -rrb- , sometimes credited as sue blu , is an american voice actress , voice director and casting director in american and canadian cinema and television .}\\
		\multicolumn{2}{p{.93\linewidth}}{\textbf{PB\&L}: susan maria blu -lrb- born july 12 , 1948 \textcolor{red}{\underline{-rrb-}} \textcolor{red}{\underline{,}} sometimes \textcolor{red}{\underline{credited}} \textcolor{red}{\underline{as}} sue blu \textcolor{red}{\underline{,}} \textcolor{red}{\underline{is}} \textcolor{red}{\underline{an}} american voice actress \textcolor{red}{\underline{,}} voice director \textcolor{red}{\underline{and}} casting director \textcolor{red}{\underline{in}} american \textcolor{red}{\underline{and}} \textcolor{red}{\underline{canadian}} cinema \textcolor{red}{\underline{and}} television .}\\
		\multicolumn{2}{p{.93\linewidth}}{\textbf{Ours}: susan maria blu -lrb- born july 12 , 1948 -rrb- \textcolor{red}{\underline{,}} \textcolor{red}{\underline{sometimes}} \textcolor{red}{\underline{credited}} \textcolor{red}{\underline{as}} \textcolor{red}{\underline{sue}} \textcolor{red}{\underline{blu}} \textcolor{red}{\underline{,}} is an american voice actress , voice director and casting director \textcolor{red}{\underline{in}} \textcolor{red}{\underline{american}} \textcolor{red}{\underline{and}} \textcolor{red}{\underline{canadian}} \textcolor{red}{\underline{cinema}} \textcolor{red}{\underline{and}} \textcolor{red}{\underline{television}} .}\\\\
		\toprule
		\textsc{key}   & \textsc{value} \\
		\midrule
		name           & patricia flores fuentes               \\
		birth\_date    & 25 july 1977 \\
		birth\_place   & state of mexico , mexico              \\
		occupation     & politician \\
		nationality    & mexican \\
		article\_title & patricia flores fuentes               \\
		\bottomrule
		\\
		\multicolumn{2}{p{.95\linewidth}}{\textbf{Ref.}: patricia flores fuentes -lrb- born 25 july 1977 -rrb- is a mexican politician affiliated to the national action party .}\\
		\multicolumn{2}{p{.95\linewidth}}{\textbf{PB\&L}: patricia flores fuentes \textcolor{red}{\underline{-lrb-}} \textcolor{red}{\underline{born}} 25 july 1977 \textcolor{red}{\underline{-rrb-}} \textcolor{red}{\underline{is}} \textcolor{red}{\underline{a}} mexican politician \textcolor{red}{\underline{affiliated}} \textcolor{red}{\underline{to}} \textcolor{red}{\underline{the}} \textcolor{red}{\underline{national}}
		\textcolor{red}{\underline{action}} party \textcolor{red}{\underline{.}}}\\
		\multicolumn{2}{p{.95\linewidth}}{\textbf{Ours}: patricia flores fuentes -lrb- born 25 july 1977 -rrb- is a mexican politician \textcolor{red}{\underline{affiliated}} \textcolor{red}{\underline{to}} \textcolor{red}{\underline{the}} \textcolor{red}{\underline{national}} \textcolor{red}{\underline{action}} \textcolor{red}{\underline{party}} .}\\\\
		\toprule
		\textsc{key}   & \textsc{value} \\
		\midrule
		name           & ate faber \\
		birth\_date    & 19 march 1894 \\
		birth\_place   & leeuwarden , netherlands              \\
		death\_date    & 19 march 1962 \\
		death\_place   & zutphen , netherlands                 \\
		sport          & fencing \\
		article\_title & ate faber \\
		\bottomrule
		\\
		\multicolumn{2}{p{.95\linewidth}}{\textbf{Ref.}: ate faber -lrb- 19 march 1894 -- 19 march 1962 -rrb- was a dutch fencer .}\\
		\multicolumn{2}{p{.95\linewidth}}{\textbf{PB\&L}: \textcolor{red}{\underline{ate}} faber \textcolor{red}{\underline{-lrb-}} \textcolor{red}{\underline{19}} march 1894 \textcolor{red}{\underline{--}} 19 march 1962 \textcolor{red}{\underline{-rrb-}} \textcolor{red}{\underline{was}} a \textcolor{red}{\underline{dutch}} fencer \textcolor{red}{\underline{.}}}\\
		\multicolumn{2}{p{.95\linewidth}}{\textbf{Ours}: ate faber -lrb- 19 march 1894 -- 19 march 1962 -rrb- was a dutch fencer .}\\\\
	\end{tabularx}
	\caption{Hallucinated words according either to our scoring procedure or to the method proposed by \citet{PerezBeltrachini2018bootstrap}.}
	\label{tab:ex_scoring1}
\end{table*}

\begin{table*}
	\centering
	\ttfamily
	\normalsize
	\begin{tabularx}{\linewidth}{lX}
		\toprule
		\textsc{key}   & \textsc{value} \\
		\midrule
		name           & alex wilmot sitwell \\
		birth\_date    & 16 march 1961 \\
		birth\_place   & uk \\
		occupation     & president , europe and emerging markets -lrb- ex-asia -rrb- of bank of america merrill lynch               \\
		article\_title & alex wilmot-sitwell \\
		\bottomrule
		\\
		\multicolumn{2}{p{.9\linewidth}}{\textbf{Ref.}: alex wilmot-sitwell heads bank of america merrill lynch 's businesses across europe and emerging markets excluding asia .}\\
		\multicolumn{2}{p{.9\linewidth}}{\textbf{PB\&L}: alex \textcolor{red}{\underline{wilmot-sitwell}} \textcolor{red}{\underline{heads}} \textcolor{red}{\underline{bank}} \textcolor{red}{\underline{of}} \textcolor{red}{\underline{america}} \textcolor{red}{\underline{merrill}} \textcolor{red}{\underline{lynch}} \textcolor{red}{\underline{'s}} \textcolor{red}{\underline{businesses}} \textcolor{red}{\underline{across}} \textcolor{red}{\underline{europe}} \textcolor{red}{\underline{and}} \textcolor{red}{\underline{emerging}} \textcolor{red}{\underline{markets}} \textcolor{red}{\underline{excluding}} \textcolor{red}{\underline{asia}} \textcolor{red}{\underline{.}}}\\
		\multicolumn{2}{p{.9\linewidth}}{\textbf{Ours}: alex wilmot-sitwell heads bank of america merrill lynch 's businesses across europe and emerging markets \textcolor{red}{\underline{excluding}} \textcolor{red}{\underline{asia}} .}\\\\
		\toprule
		\textsc{key}   & \textsc{value} \\
		\midrule
		name           & ryan moore \\
		spouse         & nichole olson -lrb- m. 2011 -rrb- \\
		children       & tucker \\
		college        & unlv \\
		yearpro        & 2005 \\
		tour           & pga tour \\
		prowins        & 4 \\
		pgawins        & 4 \\
		masters        & t12 2015 \\
		usopen         & t10 2009 \\
		open           & t10 2009 \\
		pga            & t9 2006 \\
		article\_title & ryan moore -lrb- golfer -rrb- \\
		\bottomrule
		\\
		\multicolumn{2}{p{.9\linewidth}}{\textbf{Ref.}: ryan david moore -lrb- born december 5 , 1982 -rrb- is an american professional golfer , currently playing on the pga tour .}\\
		\multicolumn{2}{p{.9\linewidth}}{\textbf{PB\&L}: ryan \textcolor{red}{\underline{david}} moore \textcolor{red}{\underline{-lrb-}} \textcolor{red}{\underline{born}} \textcolor{red}{\underline{december}} \textcolor{red}{\underline{5}} \textcolor{red}{\underline{,}} \textcolor{red}{\underline{1982}} \textcolor{red}{\underline{-rrb-}} \textcolor{red}{\underline{is}} \textcolor{red}{\underline{an}} \textcolor{red}{\underline{american}} \textcolor{red}{\underline{professional}} \textcolor{red}{\underline{golfer}} \textcolor{red}{\underline{,}} \textcolor{red}{\underline{currently}} \textcolor{red}{\underline{playing}} \textcolor{red}{\underline{on}} \textcolor{red}{\underline{the}} \textcolor{red}{\underline{pga}} \textcolor{red}{\underline{tour}} \textcolor{red}{\underline{.}}}\\
		\multicolumn{2}{p{.9\linewidth}}{\textbf{Ours}: ryan david moore \textcolor{red}{\underline{-lrb-}} \textcolor{red}{\underline{born}} \textcolor{red}{\underline{december}} \textcolor{red}{\underline{5}} \textcolor{red}{\underline{,}} \textcolor{red}{\underline{1982}} \textcolor{red}{\underline{-rrb-}} is an american professional golfer , currently playing on the pga tour .}\\\\
	\end{tabularx}

	\caption{Hallucinated words according either to our scoring procedure or to the method proposed by \citet{PerezBeltrachini2018bootstrap}.}
	\label{tab:ex_scoring2}
\end{table*}

In the subsequent tables, some WikiBio (\ref{tab:ex_output_first} to \ref{tab:ex_output_last}) and ToTTo (\ref{tab:ex_output_totto_first} to \ref{tab:ex_output_totto_last}) inputs are shown, coupled with the corresponding sentences, either as found in the dataset, or as generated by our models and baselines.

\begin{table*}\centering\small
\begin{tabular}{ll}
\toprule
    \texttt{title} & \texttt{prince of noër} \\
    \texttt{name} & \texttt{prince frederick} \\
    \texttt{image} & \texttt{prinsen af noer.jpg} \\
    \texttt{image\_size} & \texttt{200px} \\
    \texttt{spouse} & \texttt{countess henriette of danneskjold-samsøe mary esther lee} \\
    \texttt{issue} & \texttt{prince frederick , count of noer prince christian louise ,} \\& \texttt{princess michael vlangali-handjeri princess marie} \\
    \texttt{house} & \texttt{house ofschleswig-holstein-sonderburg-augustenburg} \\
    \texttt{father} & \texttt{frederick christian ii , duke of} \\& \texttt{schleswig-holstein-sonderburg-augustenburg} \\
    \texttt{mother} & \texttt{princess louise auguste of denmark} \\
    \texttt{birth\_date} & \texttt{23 august 1800} \\
    \texttt{birth\_place} & \texttt{kiel} \\
    \texttt{death\_date} & \texttt{2 july 1865} \\
    \texttt{death\_place} & \texttt{beirut} \\
    \texttt{article\_title} & \texttt{prince frederick of schleswig-holstein-sonderburg-augustenburg} \\
\end{tabular}\\
\begin{tabularx}{\textwidth}{lX}
\toprule
    Gold          & prince frederick emil august of schleswig-holstein-sonderburg-augustenburg ( kiel , 23 august 1800 -- beirut , 2 july 1865 ) , usually simply known by just his first name , frederick , `` prince of noër '' , was a prince of the house of schleswig-holstein-sonderburg-augustenburg and a cadet-line descendant of the danish royal house . \\
    \texttt{stnd} & prince frederick of schleswig-holstein-sonderburg-augustenburg ( 23 august 1800 -- 2 july 1865 ) was a member of the house of schleswig-holstein-sonderburg-augustenburg . \\
    \texttt{stnd\_filtered} & prince frederick of schleswig-holstein-sonderburg-augustenburg ( 23 august 1800 -- 2 july 1865 ) was a german . \\
    \texttt{hsmm} & prince frederick of schleswig-holstein-sonderburg-augustenburg ( 23 august 1800 -- 2 july 1865 ) was a danish noblewoman . \\
    \texttt{hier} & prince frederick of schleswig-holstein-sonderburg-augustenburg ( ) ( 23 august 1800 -- 2 july 1865 ) was a german prince of the house of schleswig-holstein-sonderburg-augustenburg . \\
    \texttt{MBD[.4, .1, .5]} & prince frederick of schleswig-holstein-sonderburg-augustenburg ( ; 23 august 1800 -- 2 july 1865 ) was the son of frederick christian ii , duke of schleswig-holstein-sonderburg-augustenburg and princess louise auguste of denmark . \\
\bottomrule
\end{tabularx}
\caption{A WikiBio input table, coupled with the corresponding sentence and the models-generated outputs.}
\label{tab:ex_output_first}
\end{table*}
\begin{table*}\centering\small
\begin{tabular}{ll}
\toprule
    \texttt{name} & \texttt{godgory} \\
    \texttt{background} & \texttt{group\_or\_band} \\
    \texttt{origin} & \texttt{karlstad , sweden} \\
    \texttt{genre} & \texttt{melodic death metal doom metal gothic metal} \\
    \texttt{years\_active} & \texttt{1992 -- 2004} \\
    \texttt{label} & \texttt{nuclear blast} \\
    \texttt{current\_members} & \texttt{matte andersson erik andersson} \\
    \texttt{past\_members} & \texttt{mikael dahlqvist stefan grundel henrik lindström fredric} \\& \texttt{danielsson thomas heder} \\
    \texttt{article\_title} & \texttt{godgory} \\
\end{tabular}\\
\begin{tabularx}{\textwidth}{lX}
\toprule
    Gold          & godgory was a swedish melodic death metal or death doom metal band that was formed august 1992 by erik andersson ( drums ) and matte andersson ( vocals ) of no relation . \\
    \texttt{stnd} & godgory was a melodic death metal band from karlstad , sweden . \\
    \texttt{stnd\_filtered} & godgory was a swedish metal band . \\
    \texttt{hsmm} & godgory ( 10 august 2004 -- 4 january 2010 ) was a melodic death metal band from karlstad , sweden . \\
    \texttt{hier} & godgory was a melodic death metal band from sweden , sweden . \\
    \texttt{MBD[.4, .1, .5]} & godgory was a swedish melodic death metal band from karlstad . \\
\bottomrule
\end{tabularx}
\caption{A WikiBio input table, coupled with the corresponding sentence and the models-generated outputs.}
\end{table*}

\begin{table*}\centering\small
\begin{tabular}{ll}
\toprule
    \texttt{name} & \texttt{christy mihos} \\
    \texttt{image} & \texttt{christy2006wp.jpeg} \\
    \texttt{px|office} & \texttt{massachusetts turnpike commission member} \\
    \texttt{term\_start} & \texttt{july 1 , 1999} \\
    \texttt{term\_end} & \texttt{july 1 , 2004} \\
    \texttt{successor} & \texttt{daniel grabauskas} \\
    \texttt{birth\_date} & \texttt{13 june 1949} \\
    \texttt{birth\_place} & \texttt{brockton , massachusetts} \\
    \texttt{nationality} & \texttt{american} \\
    \texttt{occupation} & \texttt{convenience store franchise owner} \\
    \texttt{spouse} & \texttt{andrea mihos ( 1974-2013 ; divorce )} \\
    \texttt{religion} & \texttt{greek orthodox} \\
    \texttt{alma\_mater} & \texttt{stonehill college} \\
    \texttt{party} & \texttt{republican} \\
    \texttt{article\_title} & \texttt{christy mihos} \\
\end{tabular}\\
\begin{tabularx}{\textwidth}{lX}
\toprule
    Gold          & christy peter mihos is an american politician and businessman from massachusetts . \\
    \texttt{stnd} & christy mihos ( born june 13 , 1949 in brockton , massachusetts ) is a former republican member of the massachusetts house of representatives . \\
    \texttt{stnd\_filtered} & christy mihos ( born june 13 , 1949 in brockton , massachusetts ) is an american politician . \\
    \texttt{hsmm} & christy mihos ( born june 13 , 1949 ) is an american politician from the state of massachusetts . \\
    \texttt{hier} & christy mihos ( born june 13 , 1949 ) is an american politician who served as the massachusetts state representative from 1999 to 2004 . \\
    \texttt{MBD[.4, .1, .5]} & christy mihos ( born june 13 , 1949 , in brockton , massachusetts ) is a former member of the massachusetts turnpike commission . \\
\bottomrule
\end{tabularx}
\caption{A WikiBio input table, coupled with the corresponding sentence and the models-generated outputs.}
\end{table*}

\begin{table*}\centering\small
\begin{tabular}{ll}
\toprule
    \texttt{name} & \texttt{duncan eagleson} \\
    \texttt{nationality} & \texttt{american} \\
    \texttt{area} & \texttt{painter , writer , penciller , colorist , sculptor , designer} \\
    \texttt{article\_title} & \texttt{duncan eagleson} \\
\end{tabular}\\
\begin{tabularx}{\textwidth}{lX}
\toprule
    Gold          & duncan eagleson is an american self-trained painter and former graffiti artist . \\
    \texttt{stnd} & duncan eagleson is an american painter , writer , and penciller . \\
    \texttt{stnd\_filtered} & duncan eagleson is an american painter . \\
    \texttt{hsmm} & duncan eagleson ( born 1975 ) is an american comic book painter and writer . \\
    \texttt{hier} & duncan eagleson is an american painter , illustrator , and designer . \\
    \texttt{MBD[.4, .1, .5]} & duncan eagleson is an american painter , writer , and sculptor . \\
\bottomrule
\end{tabularx}
\caption{A WikiBio input table, coupled with the corresponding sentence and the models-generated outputs.}
\end{table*}

\begin{table*}\centering\small
\begin{tabular}{ll}
\toprule
    \texttt{name} & \texttt{gerald warner brace} \\
    \texttt{imagesize} & \texttt{208px} \\
    \texttt{birth\_date} & \texttt{24 september 1901} \\
    \texttt{birth\_place} & \texttt{islip , long island , suffolk county , new york} \\
    \texttt{death\_date} & \texttt{20 july 1978} \\
    \texttt{death\_place} & \texttt{blue hill , maine} \\
    \texttt{occupation} & \texttt{novelist , writer , educator , sailor , boat builder} \\
    \texttt{nationality} & \texttt{american} \\
    \texttt{genre} & \texttt{fiction , non-fiction} \\
    \texttt{article\_title} & \texttt{gerald warner brace} \\
\end{tabular}\\
\begin{tabularx}{\textwidth}{lX}
\toprule
    Gold          & gerald warner brace ( september 24 , 1901 -- july 20 , 1978 ) was an american novelist , writer , educator , sailor and boat builder . \\
    \texttt{stnd} & gerald warner brace ( september 24 , 1901 -- july 20 , 1978 ) was an american novelist , writer , and boat builder . \\
    \texttt{stnd\_filtered} & gerald warner brace ( september 24 , 1901 -- july 20 , 1978 ) was an american novelist . \\
    \texttt{hsmm} & gerald warner brace ( september 24 , 1901 -- july 20 , 1978 ) was an american novelist and writer . \\
    \texttt{hier} & gerald warner brace ( september 24 , 1901 -- july 20 , 1978 ) was an american novelist , short story writer , educator , and sailor . \\
    \texttt{MBD[.4, .1, .5]} & gerald warner brace ( september 24 , 1901 -- july 20 , 1978 ) was an american author , educator , sailor , and boat builder . \\
\bottomrule
\end{tabularx}
\caption{A WikiBio input table, coupled with the corresponding sentence and the models-generated outputs.}
\end{table*}

\begin{table*}\centering\small
\begin{tabular}{ll}
\toprule
    \texttt{name} & \texttt{robert b. murrett} \\
    \texttt{image} & \texttt{robertbmurrett.jpg} \\
    \texttt{office} & \texttt{4th director of the national geospatial-intelligence agency} \\& \texttt{director of the office of naval intelligence} \\
    \texttt{president} & \texttt{george w. bush barack obama george w. bush} \\
    \texttt{term\_start} & \texttt{2006 2005} \\
    \texttt{term\_end} & \texttt{2010 2006} \\
    \texttt{predecessor} & \texttt{james r. clapper richard b. porterfield} \\
    \texttt{successor} & \texttt{letitia long tony l. cothron} \\
    \texttt{alma\_mater} & \texttt{university at buffalo georgetown university joint military} \\& \texttt{intelligence college} \\
    \texttt{branch} & \texttt{united states navy} \\
    \texttt{rank} & \texttt{vice admiral 20px} \\
    \texttt{article\_title} & \texttt{robert b. murrett} \\
\end{tabular}\\
\begin{tabularx}{\textwidth}{lX}
\toprule
    Gold          & vice admiral robert b. murrett was the fourth director of the national geospatial-intelligence agency , from 7 july 2006 through july 2010 . \\
    \texttt{stnd} & robert b. murrett is a retired vice admiral of the united states navy . \\
    \texttt{stnd\_filtered} & robert b. murrett is the director of the national geospatial-intelligence agency . \\
    \texttt{hsmm} & robert b. `` bob '' murrett ( born 1956 ) is an american naval officer and the director . \\
    \texttt{hier} & robert b. murrett is a retired vice admiral in the united states navy . \\
    \texttt{MBD[.4, .1, .5]} & robert b. murrett is a vice admiral in the united states navy . \\
\bottomrule
\end{tabularx}
\caption{A WikiBio input table, coupled with the corresponding sentence and the models-generated outputs.}
\end{table*}
\begin{table*}\centering\small
\begin{tabular}{ll}
\toprule
    \texttt{name} & \texttt{rosane ferreira} \\
    \texttt{image} & \texttt{deputada federal rosane ferreira.jpg} \\
    \texttt{office} & \texttt{federal deputy for state of parná} \\
    \texttt{term\_start} & \texttt{1 february 2011} \\
    \texttt{term\_end} & \texttt{actual} \\
    \texttt{president} & \texttt{dilma rousseff} \\
    \texttt{order} & \texttt{federal deputy for the state of roraima} \\
    \texttt{birth\_date} & \texttt{31 july 1963} \\
    \texttt{birth\_place} & \texttt{clevelândia , parná , brazil} \\
    \texttt{dead} & \texttt{alive} \\
    \texttt{nationality} & \texttt{brazilian} \\
    \texttt{party} & \texttt{green party ( brazil )} \\
    \texttt{article\_title} & \texttt{rosane ferreira} \\
\end{tabular}\\
\begin{tabularx}{\textwidth}{lX}
\toprule
    Gold          & rosane ferreira ( cleusa rosane ribas ferreira , born clevelândia , paraná , july 31 , 1963 ) , is a nurse and a brazilian politician . \\
    \texttt{stnd} & rosane ferreira ( born 31 july 1963 in clevelândia , parná ) is a brazilian politician . \\
    \texttt{stnd\_filtered} & rosane ferreira ( born 31 july 1963 ) is a brazilian politician . \\
    \texttt{hsmm} & rosane ferreira ( born july 31 , 1963 ) is a brazilian politician and the federal deputy . \\
    \texttt{hier} & rosane ferreira ( born 31 july 1963 ) is a brazilian politician and the current federal deputy for the state of roraima . \\
    \texttt{MBD[.4, .1, .5]} & rosane ferreira ( born 31 july 1963 in clevelândia , parná , brazil ) is a brazilian politician . \\
\bottomrule
\end{tabularx}
\caption{A WikiBio input table, coupled with the corresponding sentence and the models-generated outputs.}
\end{table*}

\begin{table*}\centering\small
\begin{tabular}{ll}
\toprule
    \texttt{name} & \texttt{polina miller} \\
    \texttt{gender} & \texttt{f} \\
    \texttt{birth\_date} & \texttt{23 november 1988} \\
    \texttt{birth\_place} & \texttt{saint petersburg , russian sfsr , soviet union} \\
    \texttt{discipline} & \texttt{wag} \\
    \texttt{article\_title} & \texttt{polina miller} \\
\end{tabular}\\
\begin{tabularx}{\textwidth}{lX}
\toprule
    Gold          & polina miller ( , born november 23 , 1988 in saint petersburg ) is a russian gymnast . \\
    \texttt{stnd} & polina miller ( born november 23 , 1988 ) is a russian artistic gymnast . \\
    \texttt{stnd\_filtered} & polina miller ( born november 23 , 1988 ) is a . \\
    \texttt{hsmm} & polina miller ( born 23 november 1988 in saint petersburg ) is a russian artistic gymnast . \\
    \texttt{hier} & polina miller ( born 23 november 1988 ) is a russian rhythmic gymnast . \\
    \texttt{MBD[.4, .1, .5]} & polina miller ( born 23 november 1988 in saint petersburg , russian sfsr , soviet union ) is a russian gymnast . \\
\bottomrule
\end{tabularx}
\caption{A WikiBio input table, coupled with the corresponding sentence and the models-generated outputs.}
\end{table*}

\begin{table*}\centering\small
\begin{tabular}{ll}
\toprule
    \texttt{name} & \texttt{pat burke} \\
    \texttt{irish} & \texttt{pádraig de búrca} \\
    \texttt{sport} & \texttt{gaelic football} \\
    \texttt{code} & \texttt{football} \\
    \texttt{county} & \texttt{dublin} \\
    \texttt{province} & \texttt{leinster} \\
    \texttt{clposition} & \texttt{corner forward} \\
    \texttt{club} & \texttt{kilmacud crokes} \\
    \texttt{clubs} & \texttt{kilmacud crokes} \\
    \texttt{counties} & \texttt{dublin} \\
    \texttt{icprovince} & \texttt{1} \\
    \texttt{birth\_place} & \texttt{dublin , ireland} \\
    \texttt{article\_title} & \texttt{pat burke ( gaelic footballer )} \\
\end{tabular}\\
\begin{tabularx}{\textwidth}{lX}
\toprule
    Gold          & pat burke is an irish gaelic footballer who plays for dublin and kilmacud crokes . \\
    \texttt{stnd} & pat burke is a gaelic footballer from dublin , ireland . \\
    \texttt{stnd\_filtered} & pat burke is a gaelic footballer for dublin . \\
    \texttt{hsmm} & pat burke ( born in dublin ) is a former irish gaelic footballer who played as a gaelic footballer . \\
    \texttt{hier} & pat burke is a former gaelic footballer for dublin . \\
    \texttt{MBD[.4, .1, .5]} & pat burke is a gaelic footballer from county dublin . \\
\bottomrule
\end{tabularx}
\caption{A WikiBio input table, coupled with the corresponding sentence and the models-generated outputs.}
\end{table*}

\begin{table*}\centering\small
\begin{tabular}{ll}
\toprule
    \texttt{name} & \texttt{odiakes} \\
    \texttt{background} & \texttt{non\_vocal\_instrumentalist} \\
    \texttt{birth\_date} & \texttt{march 22} \\
    \texttt{origin} & \texttt{tokyo , japan} \\
    \texttt{instrument} & \texttt{keyboard , synthesizer} \\
    \texttt{genre} & \texttt{j-pop , techno} \\
    \texttt{occupation} & \texttt{composer} \\
    \texttt{years\_active} & \texttt{1998 -- present} \\
    \texttt{article\_title} & \texttt{odiakes} \\
\end{tabular}\\
\begin{tabularx}{\textwidth}{lX}
\toprule
    Gold          & odiakes ( born march 22 ) is a japanese composer from tokyo , japan who has worked for a variety of visual novel companies . \\
    \texttt{stnd} & , better known by his stage name odiakes , is a japanese composer . \\
    \texttt{stnd\_filtered} & odiakes is a japanese composer . \\
    \texttt{hsmm} & odiakes `` odiakes '' ( born march 22 ) is a japanese composer . \\
    \texttt{hier} & composer ( born march 22 ) is a japanese j-pop player . \\
    \texttt{MBD[.4, .1, .5]} & odiakes ( born march 22 in tokyo , japan ) is a japanese composer . \\
\bottomrule
\end{tabularx}
\caption{A WikiBio input table, coupled with the corresponding sentence and the models-generated outputs.}
\label{tab:ex_output_last}
\end{table*}

\begin{table*}\centering\small
\begin{tabular}{ll}
\toprule
    \texttt{Club} & \texttt{Istiklol} \\
    \texttt{Season} & \texttt{2015} \\
    \texttt{League} & \texttt{Tajik League} \\
    \texttt{page\_title} & \texttt{Parvizdzhon Umarbayev} \\
    \texttt{section\_title} & \texttt{Club} \\
    \texttt{section\_text} & \texttt{As of match played 29 July 2018} \\
\end{tabular}\\
\begin{tabularx}{\textwidth}{lX}
\toprule
    Gold          & In 2015 , Umarbayev signed for Tajik League FC Istiklol . \\
    \texttt{stnd} & Umarbayev joined Tajik League side Istiklol in 2015 . \\
    \texttt{stnd\_filtered} & 2015 , he joined Istiklol in the Tajik League . \\
    \texttt{hal\_{WO}} & Parvizdzhon joined Tajik League club Istiklol in 2015 . \\
    \texttt{MBD[.4, .1, .5]} & Umarbayev signed with Istiklol ahead of the 2015 Tajik League season . \\
\bottomrule
\end{tabularx}
\caption{A ToTTo input table, coupled with the corresponding sentence and the models-generated outputs.}
\label{tab:ex_output_totto_first}
\end{table*}

\begin{table*}\centering\small
\begin{tabular}{ll}
\toprule
    \texttt{Rank} & \texttt{5} \\
    \texttt{Island} & \texttt{Hulhumeedhoo} \\
    \texttt{page\_title} & \texttt{List of islands of the Maldives} \\
    \texttt{section\_title} & \texttt{Islands by area size} \\
    \texttt{section\_text} & \texttt{This list ranks the top 10 islands of the Maldives by area .} \\
    &\texttt{Some islands in the Maldives , although geographically one}\\
    &\texttt{island , are divided into two administrative islands ( for}\\
    &\texttt{example , Gan and Maandhoo in Laamu Atoll ) .} \\
\end{tabular}\\
\begin{tabularx}{\textwidth}{lX}
\toprule
    Gold          & Hulhumeedhoo is the 5th largest island in Maldives . \\
    \texttt{stnd} & It has a area of Hulhumeedhoo km² ( Islands sq mi ) . \\
    \texttt{stnd\_filtered} & is the fourth of the Maldives in Maldives . \\
    \texttt{hal\_{WO}} & Hulhumeedhoo is the largest islands of the Maldives by area size . \\
    \texttt{MBD[.4, .1, .5]} & Hulhumeedhoo is the fifth largest island by area size . \\
\bottomrule
\end{tabularx}
\caption{A ToTTo input table, coupled with the corresponding sentence and the models-generated outputs.}
\end{table*}

\begin{table*}\centering\small
\begin{tabular}{ll}
\toprule
    \texttt{Single} & \texttt{24.7 ( Twenty-Four Seven )} \\
    \texttt{page\_title} & \texttt{Singular ( band )} \\
    \texttt{section\_title} & \texttt{2010} \\
\end{tabular}\\
\begin{tabularx}{\textwidth}{lX}
\toprule
    Gold          & In 2010 , Singular released its first single , `` 24.7 ( Twenty-Four Seven ) '' . \\
    \texttt{stnd} & The first single , 24.7 ( Twenty-Four Seven ) , was released in 2010 . \\
    \texttt{stnd\_filtered} & The band won the 24.7 ( Twenty-Four Seven ) . \\
    \texttt{hal\_{WO}} & 24.7 ( Twenty-Four Seven ) . \\
    \texttt{MBD[.4, .1, .5]} & Singular released their first album , 24.7 ( Twenty-Four Seven ) . \\
\bottomrule
\end{tabularx}
\caption{A ToTTo input table, coupled with the corresponding sentence and the models-generated outputs.}
\label{tab:ex_output_totto_last}
\end{table*}

\end{document}